\documentclass[journal]{IEEEtran}

\usepackage{times}
\usepackage{helvet}
\usepackage{courier}

\usepackage{comment}
\usepackage{graphicx}
\usepackage{amsmath}
\usepackage{mathrsfs}
\usepackage{scrextend}
\usepackage{url}
\usepackage{bm}
\usepackage{amssymb}
\usepackage{color}
\usepackage{flushend}
\usepackage{multirow}
\usepackage{cite}
\usepackage{subcaption}

\newcommand{\M}[1]{\textcolor{black}{{}#1}}

\newcommand{\R}[1]{\textcolor{black}{{}#1}}
\newcommand{\RR}[1]{\textcolor{black}{{}#1}}
\begin{document}

\title{Modality Compensation Network: \\ Cross-Modal Adaptation for Action Recognition}

\author{Sijie~Song, {\it Student Member, IEEE},
        Jiaying~Liu, {\it Senior Member, IEEE}, \\
        Yanghao~Li,
        Zongming Guo {\it Member, IEEE}
        \thanks{This work was supported by National Natural Science Foundation of China under contract No. 61772043,  Beijing Natural Science Foundation under contract No. 4192025, Microsoft Research Asia (FY19-Research-Sponsorship-115) and Peking University Tencent Rhino Bird Innovation Fund. (\emph{Corresponding author: Jiaying Liu.})}
        \thanks{The authors are with Wangxuan Institute of Computer Technology, Peking University, Beijing, 100080, China, e-mail: \{ssj940920, liujiaying, lyttonhao, guozongming\}@pku.edu.cn. }
}

\maketitle

\begin{abstract}

With the prevalence of RGB-D cameras, multi-modal video data have become more available for human action recognition. One main challenge for this task lies in how to effectively leverage their complementary information. In this work, we propose a Modality Compensation Network (MCN) to explore the relationships of different modalities, and boost the representations for human action recognition. \R{We regard RGB/optical flow videos as \emph{source} modalities, skeletons as \emph{auxiliary} modality. Our goal is to extract more discriminative features from source modalities, with the help of auxiliary modality. Built on deep Convolutional Neural Networks (CNN) and Long Short Term Memory (LSTM) networks, our model bridges data from source and auxiliary modalities by a modality adaptation block to achieve adaptive representation learning, that the network learns to compensate for the loss of skeletons at test time and even at training time. We explore multiple adaptation schemes to narrow the distance between source and auxiliary modal distributions from different levels, according to the alignment of source and auxiliary data in training.} In addition, skeletons are only required in the training phase. Our model is able to improve the recognition performance with source data when testing. Experimental results reveal that MCN outperforms state-of-the-art approaches on four widely-used action recognition benchmarks.
\end{abstract}


\IEEEpeerreviewmaketitle

\section{Introduction}
\label{sec:introduction}

\IEEEPARstart{R}{ecognition} of human actions is quite an important yet fundamental task in video analytics. \R{Human action recognition technology is required for a wide range of applications}, such as intelligent video surveillance, human-computer interaction and robotics. However, action recognition remains a challenging problem due to heterogeneous backgrounds, subtle inter-class differentiation and large intra-class variations. The key to the success of this task is enhancing the feature discriminations of various actions.

In the past decades, much work has been extensively studied on human action recognition in RGB videos~\cite{CVIU11SurveyAction}. Besides the methods relying on hand-crafted features~\cite{wang2011action}, the two-stream architecture based on deep networks is very popular and successful in utilizing visual and temporal cues~\cite{simonyan2014two}. The spatial and temporal streams are trained independently and the final results are generated from the combination of each stream. Due to the strong performance of the two-stream architecture in action recognition, many works~\cite{wang2016temporal,feichtenhofer2017spatiotemporal,feichtenhofer2016convolutional} following this line focus more on improving feature discriminations, and modelling the correlations between the spatial and temporal structures. Our starting point is the notice of the following difficulties in the task of action recognition based on the two-stream architecture:

\begin{itemize}
  \item With limited training samples, it is hard for the network to extract invariant features and cover all kinds of possible action variations, such as the differences in viewpoints and scales.

  \item For the spatial stream, \R{the extracted features can be disturbed by cluttered backgrounds, which leads to the mixture of human shapes and backgrounds.}

  \item For the temporal stream, \R{the background motions will introduce noises in the temporal features and disturb the learning process.}

\end{itemize}

To address the above issues, we are inspired to seek other cues for more effective and robust feature learning. With the recent advent of low-cost depth cameras like Microsoft Kinect~\cite{zhang2012microsoft} and the maturity of pose estimation technologies~\cite{shotton2013real}, there is an increasing amount of visual data containing both color videos and 3D skeletons. These modalities provide multiple cues \R{that are complimentary, which inspires us to explore their underlying common feature spaces. For example, the skeletons are invariant to different viewpoints and backgrounds. By exploring their common feature spaces, we expect the network to learn invariant and robust features from RGB/optical flow data.}

As a high-level representation of human motion, biological observations have suggested that humans can recognize actions from skeleton data~\cite{PP73Perception}. Skeleton-based representation is robust to the variations of illumination and scales, \R{and 3D skeletons are even invariant to different viewpoints}. Besides, it is able to exclude backgrounds, provide clear motion patterns and human posture descriptions. These advantages make skeleton data as an attractive option for action recognition~\cite{zhu2015co,song2016end,shahroudy2016ntu,liu2016spatio}. Nevertheless, the lack of appearance information results in the ambiguity for some actions whenever only observing skeletons.

To overcome the lack of appearance for skeleton data, many works integrate skeletons with other modalities to utilize complementary information. Most previous works focus on feature fusion, which is borrowed from the two-stream architecture. A recent work developed a chained multi-stream approach~\cite{zolfaghari2017chained} built on Markov chain model to combine appearance, motion and pose features. Other works tend to explore the common high-level feature spaces from different modalities through cross-modal learning~\cite{shi2017learning,mahasseni2016regularizing}. The work in~\cite{mahasseni2016regularizing} is one of the early attempts. However, it only regularizes the feature learning of the RGB stream with skeletons, without taking optical flow into consideration. \R{It is still underexplored that how to use skeletons to compensate for features from optical flow data.} Besides, the modality adaptation under different scenarios where the modalities are aligned or misaligned should be explored.

In this work, we demonstrate that the skeleton-based representation is a valuable supplement to color videos in action recognition. We focus on adapting skeletons into modal spaces learnt from RGB/optical flow data. We define RGB/optical flow data as source data and skeletons as auxiliary data. Our key idea is to borrow the advantages of auxiliary skeletons and learn the compensated representations from source data. For the spatial stream, guided by skeletons, the network is encouraged to extract complementary features reflecting motion and detailed appearance contexts from source data. For the temporal stream, skeletons are used to supplement the shape reference, and facilitate the motion feature learning to exclude the motion noises in the backgrounds. Our work needs auxiliary data provided in training, but only requires source data in the testing phase. \M{The main contributions of this paper can be summarized as follows:}
\begin{itemize}

  \item We propose a novel Modality Compensation Network for action recognition with adaptive representation learning, which aims to align the distributions of source and auxiliary data. Our model is able to extract compensated and more discriminative features from source data for action recognition.
  \item \M{A modality adaptation block with residual feature learning is developed to bridge data from source and auxiliary modalities. We show that the residual structure is more effective in borrowing the advantages of auxiliary data.}
  \item We explore different levels of modality adaptation schemes, including domain-, category- and sample-level, to cope with different scenarios according to the alignment of color and skeleton videos in the training data.
  \item \M{We give comprehensive analysis on each component in our model to better understand the modality adaptation. Our approach is evaluated on the NTU RGB+D dataset, the MSR 3D Daily Activity dataset,  the UCF-101 dataset and the JHMDB dataset, respectively. Experimental results show that our proposed model outperforms other state-of-the-art methods, thanks to more effective spatial and temporal feature learning.}
\end{itemize}

The remainder of the paper is organized as follows. In Section~\ref{sec:relatedwork}, we review the related works on action recognition based on RGB videos or skeletons, and multi-modal feature learning. In Section~\ref{sec:proposed}, we first give an overview of our baseline structure, and then introduce the proposed Modality Compensation Network in detail. Experiments and analysis are presented in Section~\ref{sec:experiments}. Finally, we conclude our work in Section~\ref{sec:conclusion}.

\vspace{-2mm}
\section{Related Work}
\label{sec:relatedwork}
\subsection{RGB-Based Action Recognition}

\R{Action recognition on RGB-based videos has attracted many research interests and been extensively studied in the past decades. For action recognition with RGB as inputs, a key branch is based on hand-crafted descriptors with decoupled classifiers~\cite{laptev2008learning,wang2009evaluation,wang2011action,wang2013action}, other methods try to jointly learn features and classifiers. With the advent of deep learning, neural networks are recently employed for action recognition due to its powerful ability in learning robust feature representations~\cite{simonyan2014two,karpathy2014large, yue2015beyond,tran2015learning, fernando2016discriminative, bilen2016dynamic,wang2016temporal,wang2017spatiotemporal, feichtenhofer2017spatiotemporal,feichtenhofer2016convolutional,zhang2016action,feichtenhofer2016spatiotemporal}. } The two-stream architecture~\cite{simonyan2014two} is a pioneer work to employ deep convolutional network for action recognition in videos, and has become a backbone of many other approaches~\cite{wang2016temporal,wang2017spatiotemporal, feichtenhofer2017spatiotemporal,feichtenhofer2016convolutional,feichtenhofer2016spatiotemporal,li2019temporal,jiang2018rethinking}. To address the aggregation of spatial and temporal features,~\cite{feichtenhofer2016convolutional} explored different score fusion schemes. The work in~\cite{feichtenhofer2017spatiotemporal,feichtenhofer2016spatiotemporal} utilized residual connections to allow spatiotemporal interaction between to streams.~\cite{tran2015learning} used 3D convolutional networks (C3D) to learn discriminative spatio-temporal patterns jointly. Based on C3D, VLAD3~\cite{li2016vlad3} was developed to model long-range dynamic information. Besides, the works in~\cite{fernando2016discriminative,bilen2016dynamic} performed a hierarchical rank pooling to obtain video representations, which have high capacity of capturing informative frame-based feature representations. To further extend the temporal support, great efforts have been made with recurrent neural networks, learning how to integrate dynamics overtime~\cite{yue2015beyond,donahue2015long,wu2016multi,wu2015modeling}. Towards good practice, \cite{wang2016temporal} presents a Temporal Segment Network (TSN) and provides an effective way to model long-term temporal structure, which has brought the state-of-the-art to a new stage. \R{However, the diversities in viewpoints and backgrounds make it challenging to extract discriminative features from RGB videos. In our work, we use skeleton data, which are invariant in backgrounds and scales, to encourage the network to learn more robust features from color videos.}

\subsection{Skeleton-Based Action Recognition}
Skeleton-based action recognition can be regarded as a recognition task on time series for exploring spatio-temporal patterns. Many traditional works focus on hand-crafted features, which are generally based on the geometry relationships and high order encodings~\cite{vemulapalli2014human,vemulapalli2016rolling,wang2012mining,goutsu2015motion,lv2006recognition}. Benefiting from the merits of recurrent neural network for sequential data, some works adopt recurrent neural network to explore the spatial and temporal dynamics of skeletal data~\cite{CVPR15HRNN,du2016representation,zhu2015co,liu2016spatio,song2016end,wang2017modeling,song2018spatio}. With the help of convolutional neural networks, Ke \emph{et al.}~\cite{ke2017new} proposed a new representation for 3D skeleton data, which transfers action recognition problem to the problem of image classification.

Due to lack of appearance for skeleton data, many works integrate skeletons with data from other modalities to utilize the complementary information. \R{A regularized LSTM is proposed~\cite{mahasseni2016regularizing} to constrain the RGB feature space with skeletons. But the relationship between skeletons and optical flow data is overlooked in~\cite{mahasseni2016regularizing}. Shi and Tim~\cite{shi2017learning} proposed to achieve action recognition from depth sequences by learning an RNN with privileged information from skeletons. In addition, a chained multi-stream network~\cite{zolfaghari2017chained} built on Markov chain model is developed to integrate appearance, motion and pose features. Liu \emph{et al.}~\cite{liu2018multi} jointly learned the regression and classification network with multi-modal data for action detection. Though the methods in~\cite{shi2017learning,zolfaghari2017chained,liu2018multi,song2018skeleton} illustrate that the introduction of multiple cues improve the performance for action analytics, they are limited by the strict data requirements.~\cite{shi2017learning} requires aligned skeletons and depth data in training. The models in~\cite{zolfaghari2017chained} and~\cite{liu2018multi} need aligned color videos and 3D skeletons (or 2D poses) both in training and testing. However, aligned multi-modal data are not always reliable. It is desirable to utilize unaligned data to facilitate feature learning. In our work, we explore the relationship not only between RGB videos and skeletons, but also between optical flow and skeleton data. Besides, our model is flexible in that it can be trained with aligned or unaligned multi-modal data. And only source modalities (\emph{i.e.}, RGB and optical flow) are required at the test time.}

\subsection{Multi-Modal Feature Learning}
Inspired by that different modalities provide complementary information, cross-modal feature learning has been widely studied these days. Much work tends to unify the distributions of different feature spaces, by regularizing the representation learning with data from other modalities~\cite{wang2015mmss,wang2016learning}. This can be regarded as a domain adaptation problem, which aims to minimize the distribution between source and target domain~\cite{pan2010survey,ganin2015unsupervised}. Long \emph{et al.}~\cite{long2015learning} introduced a deep adaptation network to minimize the maximum mean discrepancy of the feature to learn transferable features in high layers of the network. However, the work in~\cite{bousmalis2016domain} claims that focusing only on the shared representation leads to the ignorance of individual characteristics. Following this line, Wang \emph{et al.}~\cite{wang2015mmss} proposed a multi-modal feature learning framework for RGB-D object recognition to learn not only modal-specific patterns but also modal-shared features. The similar idea is also employed in the task of image segmentation~\cite{wang2016learning}, in which RGB and depth data sources are correlated with multi-kernel maximum mean discrepancy to discover common and specific features.


\begin{figure*}[t] 
	\begin{center}
		\includegraphics[width=1.0\linewidth]{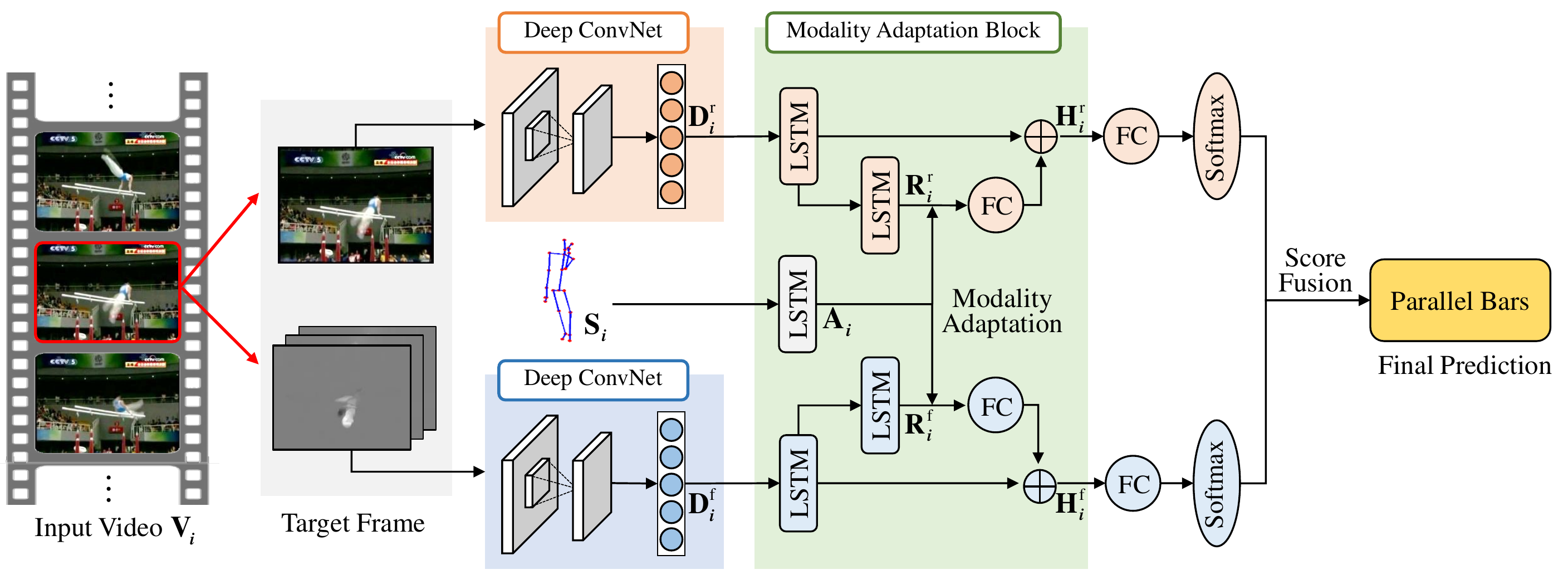}
	\end{center}
\vspace{-2mm}
	\caption{\R{The framework of Modality Compensation Network. Based on the two-stream architecture, a modality adaptation block is incorporated to compensate the feature learning of source data with the help of auxiliary data. (We use superscripts `r' and `f' to indicate RGB and flow streams, respectively, which are omitted in the text for simplicity.) Note that the LSTM with output $A_i$ is not included in the testing phase.}}
\vspace{-5mm}
	\label{fig:framework}
\end{figure*}

\section{Modality Compensation Network}
\label{sec:proposed}
We propose a modality compensation network for action recognition. Our network is designed to implicitly compensate features from source modalties with the help of auxiliary data. We propose a modality adaptation block to improve the ability of adaptation learning with skeletons for action recognition. A hierarchical modality adaptation scheme is further explored for different data alignment cases to borrow information from the auxiliary data more effectively. Fig.~\ref{fig:framework} shows the overall architecture of our network. In the following, we first introduce our baseline model, then we discuss the modality adaptation block and our hierarchical modality adaptation scheme.

\subsection{Baseline Model}
Our backbone is built on the two-stream structure~\cite{simonyan2014two} as shown in Fig.~\ref{fig:framework}: the spatial stream leverages the appearance information extracted from the RGB frames, while the temporal stream utilizes the motion contexts from the stacked horizontal and vertical optical flow data. The structure of each stream mainly consists of a CNN and an LSTM. The CNN is employed as an encoder to extract spatial features from each input frame, and the information over time is collected and integrated by the LSTM. Note that each stream is trained independently and the final results for two streams are given by score fusion. Note, we use superscripts `r' and `f' in Fig.~\ref{fig:framework} to indicate RGB and optical flow streams, respectively. The following illustrations for our proposed model are in terms of a single stream and we omit the superscripts for simplicity.

\textbf{Convolutional Neural Network.} Compared with hand-crafted features, convolutional networks have shown superiority of interpreting and summarizing images. Given the $i$-th video sequence $\mathbf{V}_i = \{\mathbf{v}_{i,t}: t = 1, ..., T\}$ within a batch, where $\mathbf{v}_{i,t}$ is the frame at time step $t$, we extract spatial features from the convolutional layers. Inspired by the pioneer work~\cite{wang2016temporal}, in experiments we employ features after \emph{global-pool} layer from the Inception with Batch Normalization (BN-Inception) network~\cite{ioffe2015batch}. Then each frame is represented by a vector $\mathbf{d}_{i,t} \in \mathbb{R}^{1024}$. We then feed the sequence of CNN features $\mathbf{D}_i = \{\mathbf{d}_{i,t}: t = 1, ..., T\}$ to the following recurrent layers.

\textbf{Long Short Term Memory Network.}
To exploit the temporal dependencies in video sequences, LSTM~\cite{Graves2012} is employed to build our network. The self-connected architecture allows LSTM to maintain and process temporal information over time. At each time step, the network can choose to read, write or reset the memory cell governed by the input gate, forget gate and output gate.

\RR{\textbf{Loss Function.} To obtain the class predictions of $\mathbf{V}_i$, the hidden states $\mathbf{H}_i = [\mathbf{h}_{i,1}, ..., \mathbf{h}_{i,T}]$ from LSTM layers is accumulated and then mapped to a 1-of-$C$ encoding vector $\mathbf{g}_i = [g_{i,1}, ..., g_{i,C}]$, representing the confidence scores of $C$ classes of actions}
\vspace{-1mm}
\begin{equation}
\vspace{-1mm}
  \mathbf{g}_i = \mathbf{W}_g \left( \frac{1}{T} \sum_{t=1}^T \mathbf{h}_{i,t}\right) + \mathbf{b}_g,
\end{equation}
\RR{where $\mathbf{W}_g$ and $\mathbf{b}_g$ are the parameters of the last FC layer before \emph{softmax}. The predicted probability of the video $\mathbf{V}_i$ being the $c$-th class is then normalized to $p(c|\mathbf{V}_i) = e^{g_{i,c}} /\sum_{j=1}^C e^{g_{i,j}} (c = 1, ..., C)$. Our goal for each stream is to minimize the objective function as}
\vspace{-1mm}
\begin{equation}
\label{equ:baseline_loss}
\vspace{-1mm}
  \mathcal{L} = -\sum_{i=1}^n \sum_{c=1}^C l_{i,c}\log p(c|\mathbf{V}_i),
\end{equation}
\RR{where $l_{i,c}\!=\!1$ if the $i$-th video sequence belongs to the $c$-th class, and 0 otherwise.}

\RR{For the final prediction, we obtain the probability for being the $c$-th class from each stream by score fusion, which can be formulated as}
\begin{equation}
\label{equ:score_fusion}
  p^*(c|\mathbf{V}_i) = \sum_{m\in\{\mathrm{r},\mathrm{f}\}} w^m p^m(c|\mathbf{V}_i),
\end{equation}
\RR{where the superscript $m$ indicates which stream the score is from. $w^m$ denotes the fusion weight, and $\sum_{m\in\{\mathrm{r},\mathrm{f}\}}w^m = 1$. In our experiments, we set $w_m = 0.5$ in two-stream fusion.}

\vspace{-3mm}
\subsection{Adaptive Representation Learning}
In the baseline model, the inter- and intra-class variations result in the difficulties in extracting robust features to adapt to different viewpoints and scales. Besides, the cluttered backgrounds can easily introduce noises in the process of feature learning. In this work, we propose an adaptive representation learning method, for the purpose of compensating features and exploring more discriminative representations from the input frames. Thanks to the skeleton data, as a high level representation invariant to viewpoints and backgrounds, we are able to improve the feature learning of source data.

\textbf{Modality Adaptation Block.} We design a modality adaptation block to achieve adaptive representation learning. For each stream, it consists of a main network and a residual subnetwork~\cite{he2016deep} using recurrent layers. \R{The idea of residual LSTM networks has been explored in~\cite{wang2016recurrent} and~\cite{hasan2016neural}, both of which demonstrates the effectiveness of the structure. Our network is more similar to the staked LSTM~\cite{hasan2016neural}. Fig.~\ref{fig:framework} shows the detail structure of our network}. On one hand, the network is able to keep the original information from source modalities. On the other hand, the compensated features are attained from the residual block through adaptive representation learning.
The basic formulation of the residual subnetwork is defined as
\begin{equation}
\label{equ:res_block}
  \mathbf{z}^{l+1} = \mathcal{H}(\mathbf{z}^{l}, \{\mathbf{\theta}^l\}) + \mathbf{z}^{l},
\end{equation}
where $\mathbf{z}^{l}$ and $\mathbf{z}^{l+1}$ are the input and output vectors of the layer considered, and $\mathcal{H}(\cdot)$ is a nonlinear residual mapping with parameters denoted as $\mathbf{\theta}^l$. More specifically, we use an LSTM layer and an FC layer as the residual mapping function in the residual subnetwork, as shown in Fig.~\ref{fig:framework}.

The insight of our adaptive representation learning is to transform the source modal space to the auxiliary modal space, so that the advantages in the auxiliary modality can also be represented using data from source modalities, and then to achieve modality compensation. \R{Note that the modality compensation is mainly achieved by the residual path, which pushes the feature representation of source modalities be close to the auxiliary skeleton features. For the features which are present in the source modalities but absent in the auxiliary modality, we use the skip connection in the residual block to preserve them.} For a given skeleton sequence $\mathbf{S}_i = \{\mathbf{s}_{i,t}: t = 1, ..., T\}$, we encode it into feature vectors $\mathbf{A}_i = \{\mathbf{a}_{i,t}: t = 1, ..., T\}$ by a pre-trained LSTM network with parameters $\{\theta_a\}$, \M{which can be formulated as $\mathbf{A}_i = f_a(\mathbf{S}_i, \{\theta_a\})$. The features from the source modalities are denoted as $\mathbf{R}_i = \{\mathbf{r}_{i,t}: t = 1, ..., T\}$, which are extracted from the output of the LSTM layer in the residual subnetwork as shown in Fig.~\ref{fig:framework}. We formulate it as $\mathbf{R}_i = f_r(\mathbf{V}_i, \{\theta_r\})$, where $\{\theta_r\}$ denotes the parameters. With fixed $\{\theta_a\}$, our goal is to learn features from source modalities similar to the pre-defined auxiliary features through optimizing the feature learning with $\{\theta_r\}$ in source data. The optimization can be achieved by minimizing the distance between the source modal space $\bar{\mathbf{R}}$ and the auxiliary modal space $\bar{\mathbf{A}}$ as follows}
\vspace{-1mm}
\begin{equation}
\vspace{-2mm}
\label{equ:trans}
  dist\left(\mathbf{\bar{R}},\mathbf{\bar{A}}\right).
\end{equation}
\RR{And it can be integrated with the loss function defined in Eq.~(\ref{equ:baseline_loss})}
\vspace{-1mm}
\begin{equation}
\label{equ:loss}
\vspace{-1mm}
\mathcal{L} = -\sum_{i=1}^n \sum_{c=1}^C l_{i,c}\log p(c|\mathbf{V}_i) + \lambda d,
\end{equation}
\RR{where $d = dist\left(\mathbf{\bar{R}},\mathbf{\bar{A}}\right)$, and we use $\lambda$ to balance the contribution of the adaptive term. However, due to the huge gap between skeletons and color videos, it is hard to define the distance function $dist(\cdot)$ directly. Instead, we bridge the different modalities by adapting their contributions from different levels, namely, domain-level adaptation ($d_D$), category-level adaptation ($d_C$), and sample-level adaptation ($d_S$). Thus we have $d \in \{d_D, d_C, d_S\}$ which will be discussed in the next subsection. }

\vspace{-2mm}
\subsection{Different Levels of Modality Adaptation}
To calculate the distance of source and auxiliary modal spaces and achieve modality adaptation, we consider two cases here:
\begin{itemize}

  \item General case: the modalities in the training data are not aligned individually.

  \item Specific case: each color video in the training data is provided with its corresponding skeleton sequence.

\end{itemize}

To deal with the different cases, we consider different levels of modality adaptation, including domain-, category- and sample-level adaptation. We summarize the requirements for multi-modal data of each adaptation scheme in Table~\ref{table:modal_adap}.
\vspace{-2mm}
\begin{table}[htbp]
	\begin{center}
		\caption{Requirements for multi-modal data of each scheme.}

		\label{table:modal_adap}
		\begin{tabular}{c||c|c|c}
			\hline
			Level & Required Alignment & Granularity & Target Case\\
            \hline
            \hline
			Domain & Not required & General & General~\&~Specific \\
			\hline
			Category  & Category-aligned  & Mid-level & Specific\\
			\hline
			Sample  & Counterparts & Specific & Specific\\
			\hline
		\end{tabular}
	\end{center}
	\vspace{-2mm}
\end{table}

\subsubsection{Domain-Level Adaptation}
Domain-level adaptation is able to deal with the most general cases, regardless of the alignment of source and auxiliary data. Inspired by domain adaptation, we utilize Maximum Mean Discrepancy (MMD)~\cite{gretton2012kernel} to align domain-level distributions and handle this modality adaptation problem. Given two sets of samples, $X = \{\mathbf{x}_i\}_{i=1}^{m_x}$ and $Y = \{\mathbf{y}_j\}_{j=1}^{m_y}$ generated from two distributions, respectively, the squared MMD calculates the norm of the difference between embeddings of the two different distributions as
\vspace{-2mm}
\begin{align}
\label{equ:mmd}
\vspace{-1mm}
\nonumber& \text{MMD}^2[X, Y]  \\
\nonumber        = ~ &\| \mathbf{E}_x[\phi(\mathbf{x})] - \mathbf{E}_y[\phi(\mathbf{y})] \|^2\\
\nonumber		= ~&\frac{1}{m_x^2}\sum_{i=1}^{m_x}\sum_{i'=1}^{m_x}\phi(\mathbf{x}_i)^T\phi(\mathbf{x}_{i'}) +
		   \frac{1}{m_y^2}\sum_{j=1}^{m_y}\sum_{j'=1}^{m_y}\phi(\mathbf{y}_j)^T\phi(\mathbf{y}_{j'}) \\
		&   -\frac{2}{m_x m_y}\sum_{i=1}^{m_x}\sum_{j=1}^{m_y}\phi(\mathbf{x}_i)^T\phi(\mathbf{y}_{j}),
\end{align}
where $\phi$ is the explicit feature mapping function of MMD, $m_x$ and $m_y$ indicate the number of samples under the two different distributions, respectively.

For the assessment of the distance between source and auxiliary modal distributions, we consider the video-level representations to compute the MMD distance.
Given the encoded features $\mathbf{R}_i$ from video $\mathbf{V}_i$, and $\mathbf{A}_i$ from skeleton sequence $\mathbf{S}_i$, respectively, we generate the video-level feature descriptors by $\mathbf{\hat{r}}_i = \frac{1}{T}\sum_{t=1}^T{\mathbf{r}_{i,t}}$ for source data, and $\mathbf{\hat{a}}_i = \frac{1}{T}\sum_{t=1}^T{\mathbf{a}_{i,t}}$ for auxiliary data.
Applying the associated kernel function $k(\mathbf{x},\mathbf{y}) = <\phi(\mathbf{x}),\phi(\mathbf{y})>$ in Eq.~(\ref{equ:mmd}), the MMD distance between the source and auxiliary data can be estimated as

\vspace{-2mm}
\begin{equation}
\vspace{-1mm}
\label{equ:domain_adap}
\begin{aligned}
d_D =&~\frac{1}{n^2}\sum_{i=1}^{n}\sum_{i'=1}^{n}k(\mathbf{\hat{a}}_i, \mathbf{\hat{a}}_{i'}) +
		   \frac{1}{n^2}\sum_{j=1}^{n}\sum_{j'=1}^{n}k(\mathbf{\hat{r}}_j, \mathbf{\hat{r}}_{j'}) \\
	& -\frac{2}{n^2}\sum_{i=1}^{n}\sum_{j=1}^{n}k(\mathbf{\hat{a}}_i, \mathbf{\hat{r}}_j),
\end{aligned}
\end{equation}
where $n$ is the batch size, and we employ the linear kernel $k(\mathbf{x},\mathbf{y}) = \mathbf{x}^T \mathbf{y}$ in our work. In practice, for misaligned source and auxiliary data (\emph{i.e.}, RGB and skeletons from different video sources in training), Eq.~(\ref{equ:domain_adap}) is optimized within a mini-batch, to explore the underlying common spaces of two misaligned modalities.

\subsubsection{Category-Level Adaptation}
\label{sec:cat_mmd}
Taking the category information into account, the network is able to learn more specific features guided by the skeleton data from the same category, so it is easier to find the optimal. In our category-based MMD, we only consider the distributions of the source and auxiliary data sharing the same label. More specifically, with the video-level feature descriptors and a linear kernel, we use the following function to calculate category-based MMD within a mini batch with the size of $n$,
\vspace{0mm}
\begin{equation}
\label{equ:cat_adap}
\begin{aligned}
d_C =& ~\frac{1}{C}\sum_{c=1}^{C}\biggl(~\frac{1}{n_c^2}\sum_{i=1}^{n_c}\sum_{i'=1}^{n_c}k(\mathbf{\hat{a}}_{i,c}, \mathbf{\hat{a}}_{i',c}) + \\
	 & ~\frac{1}{m_c^2}\sum_{j=1}^{m_c}\sum_{j'=1}^{m_c}k(\mathbf{\hat{r}}_{j,c}, \mathbf{\hat{r}}_{j',c})
	 -\frac{2}{n_c m_c}\sum_{i=1}^{n_c}\sum_{j=1}^{m_c}k(\mathbf{\hat{a}}_{i,c}, \mathbf{\hat{r}}_{j,c})\biggr),
\end{aligned}
\vspace{0mm}
\end{equation}
where there are $C$ classes of actions in total, $n_c$ and $m_c$ denote the number of auxiliary and source modal samples of the $c$-th class within the batch (\emph{i.e}., $\sum_{c=1}^C n_c = \sum_{c=1}^C m_c = n$), $\mathbf{\hat{a}}_{i,c}$ and $\mathbf{\hat{r}}_{j,c}$ are the auxiliary and source modal features of the $c$-th class, respectively.

\subsubsection{Sample-Level Adaptation}
This scenario assumes that, each color video of the source modality is provided with its corresponding skeleton sequence.
Intuitively, we can adopt the sample-level adaptation on the granularity of each frame or each video. However, despite that each input frame is corresponding to a skeleton, optical flow images could be flat whenever actions only involve subtle actions. Therefore, it is not feasible to narrow the frame-wise distance between skeleton descriptions and features from flat optical flow images. Meanwhile, no gain is observed when narrowing frame-wise distance between RGB images and skeletons. It is probably because the noises in skeleton data affect the feature learning from RGB images. Instead, we use the video-level descriptions mentioned in Eq.~(\ref{equ:domain_adap}). Euclidean distance is employed to learn adaptive features as follows
\vspace{-1mm}
\begin{equation}
\vspace{-1mm}
\label{equ:sample_adap}
  d_S = \frac{1}{n}\sum_{i=1}^n||\hat{\mathbf{a}}_i-\hat{\mathbf{r}}_i||^2,
\end{equation}
where $\hat{\mathbf{a}}_i$ and $\hat{\mathbf{r}}_i$  are the video-level descriptors of the $i$-th sample from the auxiliary and source modalities, respectively, and $n$ is the batch size.

\section{Experiment Results}
\label{sec:experiments}
For evaluation, we conduct our experiments on the following four datasets: the NTU RGB+D dataset~\cite{shahroudy2016ntu}, the MSR 3D Daily Activity Dataset~\cite{wang2012mining}, the UCF-101 dataset~\cite{soomro2012ucf101} and the JHMDB dataset~\cite{Jhuang2013ICCV}. To further explore the effectiveness of each component in our work, the ablation analysis is given. Note that our model needs data from auxiliary modality to enable better feature learning in training but only requires data from source modalities when testing.
\vspace{-1mm}
\subsection{Datasets and Settings}
\textbf{NTU RGB+D Dataset (NTU)}~\cite{shahroudy2016ntu} is the largest action recognition dataset with individually aligned multi-modal data (\emph{i.e.}, skeletons, RGB, depth and infrared). This dataset consists of 56880 video samples with more than 4 million frames. There are 60 action types performed by 40 subjects, including interactions with pairs and individual activities. Each skeleton has 25 joints. The cross subject (CS) and cross view (CV) settings are two protocols to evaluate the performance. Because the resolution of RGB frames is $1920\times1080$, after resized to $224\times224$, the actors only hold a small place in the scene. To avoid unnecessary degradation in performance, the original RGB images are cropped to increase the resolution of subjects. To crop RGB images, after mapping skeletons to RGB, the min and max of the joint coordinates on the RGB video $(x_{min}, x_{max}, y_{min}, y_{max})$ can be calculated. Then we crop the region of $[x_{min}-250 : x_{max}+250, y_{min}-50: y_{max}+50]$, increasing the resolution of actors. We also perform data normalization~\cite{shahroudy2016ntu} on skeletons to have position and view invariance. \R{To accelerate the extraction of optical flow~\cite{wang2016temporal}, the RGB and skeleton sequences are downsampled with a stride as 5 over the temporal axis.}

\M{\textbf{MSR 3D Daily Activity Dataset (MSR)}~\cite{wang2012mining} includes 16 action categories related to daily activities, such as \emph{playing the guitar}, \emph{using vacuum cleaner}. This dataset provides skeletons (20 joints for each skeleton), RGB and depth images. The total number of video sequences is 320 with image resolution of $640\times480$. The first five actors are used for training and others for testing, which follows the cross-subject setting.}

\textbf{UCF-101 Dataset (UCF-101)}~\cite{soomro2012ucf101} consists of 101 action classes, over 13k fully-annotated action videos. Each video snippet lasts 3-10 seconds and contains 100-300 frames, with a fixed resolution of $320\times240$. This dataset is challenging due to the large variations in action categories, viewpoints and backgrounds. We follow the evaluation in~\cite{wang2016temporal} and report the mean average accuracy over the three training/testing splits. The videos are downsampled and processed in sequences of 60 frames.

\textbf{JHMDB Dataset (JHMDB)}~\cite{Jhuang2013ICCV} is composed of 928 RGB videos for 21 actions such as \emph{brushing hair}, \emph{running}, \emph{shooting ball}. Each clip contains 15-40 frames of size $320\times240$ and there are 31838 frames in total. The dataset is divided into three train/test splits, and evaluation averages the results over the three splits.

\R{\textbf{Auxiliary data sources.} The auxiliary data may come from internal or external sources. We consider different auxiliary data sources for each dataset. For the NTU dataset, we consider two settings here: (1) Using its own skeletons as the auxiliary data, since the NTU dataset is provided with aligned skeleton sequences and RGB videos. (2) Using external skeletons from PKU Multi-Modality Dataset (PKU)~\cite{liu2017pku}. There are 52 action labels in the PKU dataset, and all of them are from the NTU dataset. Thus, the two datasets are category-aligned and we are able to further evaluate the effectiveness of domain-level and category-level adaptation schemes. For the MSR dataset, there are aligned skeleton sequences and RGB videos, and its own skeletons are used as auxiliary data. However, for the UCF-101 and JHMDB datasets, there are no counterpart skeletons available. We also consider two settings here to compensate the loss of skeletons: (1) Using external 3D skeletons from the NTU (CV) dataset as their auxiliary data. (2) Using 2D poses extracted from each video with an off-the-shelf pose extractor (\emph{i.e.}, OpenPose~\cite{cao2017realtime}) as their internal auxiliary data. The summary for auxiliary data sources can be found in Table~\ref{table:param}.}

\begin{figure}[t]
	\centering
    \includegraphics[width=0.8\linewidth]{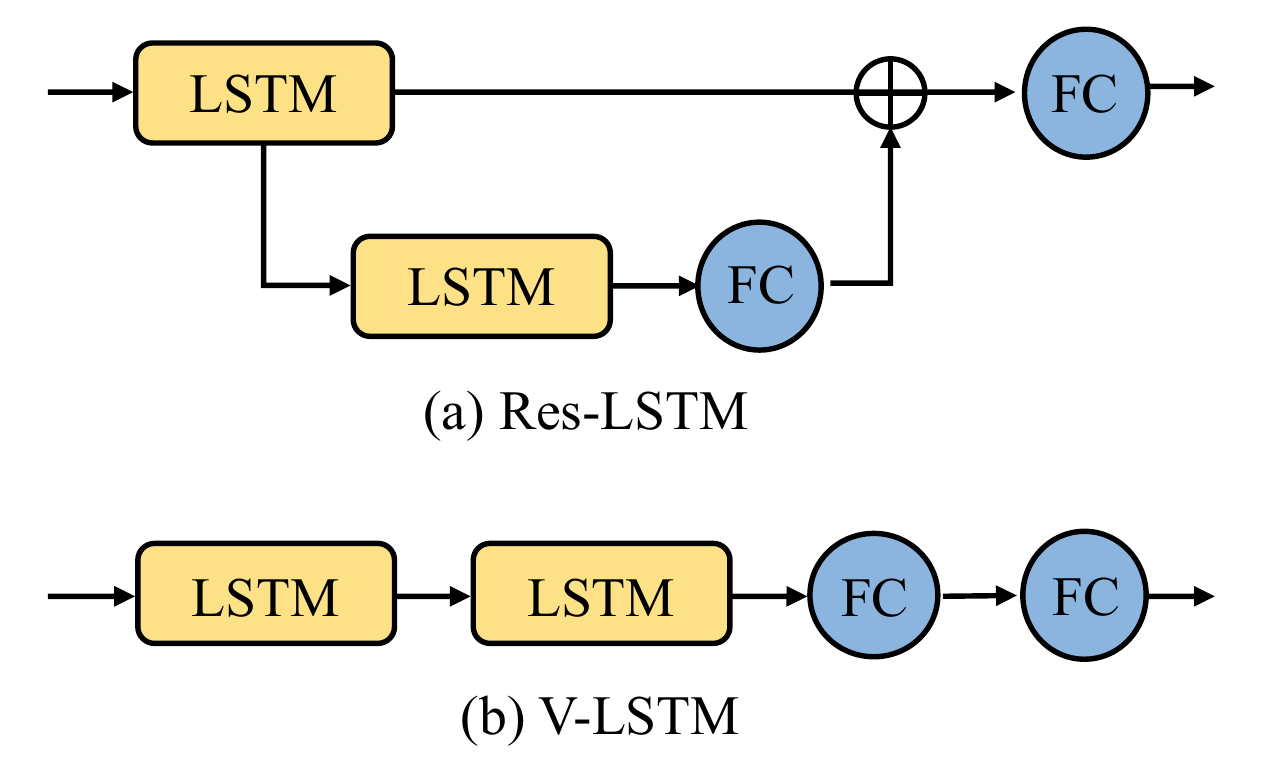}
\vspace{-2mm}
	\caption{The structures of Res-LSTM and V-LSTM, respectively. The input features are from CNN (omitted in the figure).}
\vspace{-2mm}
\label{fig:baseline}
\end{figure}

\textbf{Implementation Details.} We list the parameter settings for each dataset in Table~\ref{table:param}. For the auxiliary data, we encode the skeleton representations using a 1-layer LSTM with $N$ units, which is pretrained by the classification task. \R{For the NTU and MSR datasets, we use skeletons from the corresponding training set to pre-train the auxiliary network. For example, for NTU dataset, when under the setting of cross view (CV), the auxiliary network is trained with skeletons from the training set of NTU-CV. While under the setting of cross subject (CS), we use skeletons from the training set of NTU-CS to pre-train the auxiliary network. For UCF-101 and JHMDB dataset using external sources as auxiliary data, we simply use skeletons from the training set of NTU-CV to achieve pretraining. And when using extracted poses as auxiliary data, we use the poses from corresponding training set to pre-train the auxiliary network.} For the source data (RGB and optical flow), we pretrain the BN-Inception network in~\cite{wang2016temporal} on the four datasets, respectively. The features from the layer of \emph{global pool} in BN-Inception are extracted into a 1024-dimensional vector and then fed into the recurrent layers. We use a 2-layer LSTM network with $N$ units in each LSTM layer, including one in the residual subnetwork, as shown in Fig.~\ref{fig:baseline}(a). The number of units in the FC layer after the LSTM is also $N$. In practice, we fix the parameters for encoding skeleton representations and train the network for RGB/optical flow streams. Adam optimizer~\cite{kingma2014adam} is adopted to automatically modulate the learning rate with initial value as 0.001. Dropout~\cite{ICLR15DropoutLSTM} with a probability of 0.5 is used to mitigate overfitting. \R{For the selection of the hyper parameter $\lambda$ in Eq.~(\ref{equ:loss}), we approximate the weights per the order of magnitude for each adaptation scheme. Then we try different weights with a factor of 10 for the regularizations and choose the best weight per the performance on the validation set. Section~\ref{sec:param} further discusses the effect of $\lambda$.}

\begin{table}[t]
\fontsize{8pt}{9pt}\selectfont\centering
	\begin{center}
		\caption{Settings for each dataset.}
		\label{table:param}
        \begin{tabular}
        {l|c|c|c|c|c|c|c}
          \hline
                     & \multicolumn{2}{c|}{NTU} & MSR & \multicolumn{2}{c|}{UCF-101} & \multicolumn{2}{c}{JHMDB} \\
          \hline
          \hline
          Auxiliary data & NTU & PKU & MSR & NTU & Poses & NTU & Poses \\
          \hline
          $N$ & 1024 & 1024 & 100 &  1024 & 1024 & 100 & 100 \\
          \hline
          batch size & 256 & 256 & 16 & 192 & 192  & 16 & 16 \\
          \hline
        \end{tabular}
     \end{center}
     \vspace{-4mm}
\end{table}
\vspace{-4mm}
\subsection{Effectiveness of Modality Adaptation}
To validate the effectiveness of modality adaptation, we compare the proposed adaptation schemes on the four datasets, respectively.
We conduct experiments with the configurations targeting different cases in Table~\ref{table:modal_adap} as follows:
\begin{itemize}
  \item \textbf{Res-LSTM}: Our baseline without using modality adaptation, as shown in Fig.~\ref{fig:baseline}(a).

  \item \textbf{D-Res-LSTM}: Res-LSTM with domain-level adaptation, targeting the general and specific cases.

  \item \textbf{C-Res-LSTM}: Res-LSTM with category-level adaptation, targeting the specific case.

  \item \textbf{S-Res-LSTM}: Res-LSTM with sample-level adaptation, targeting the specific case.

  \item \textbf{J-Res-LSTM}: Res-LSTM with domain-, category-, and sample-level adaptations jointly, targeting the specific case.

\end{itemize}

Table~\ref{table:result_adap_ntu_msr} shows the results for the NTU and MSR datasets on the modalities of RGB (RGB) and optical flow (Flow), respectively. \R{For the NTU dataset, when using external auxiliary skeletons from the PKU dataset, we are able to apply domain-level and category-level adaptation schemes since they are category-aligned. Results show that both the adaptation schemes (D-Res-LSTM$^*$ and C-Res-LSTM$^*$) can improve the action recognition performance. Furthermore, the category-level adaptation outperforms domain-level adaptation. When using internal auxiliary data, it can be regarded as the specific case. We evaluate the domain-, category- and sample-level adaptation schemes, respectively. It is observed that the adaptive representation learning effectively boosts the results under all settings with internal auxiliary data. Since D-Res-LSTM and C-Res-LSTM adapt the feature learning in a coarse level, they achieve inferior results than sample-level adaptation. However, S-Res-LSTM narrows the distance between source and auxiliary modal distributions in a finer granularity. Compared with Res-LSTM, S-Res-LSTM brings about 2-3\% improvement on the NTU dataset. We also notice that the jointly adaption scheme J-Res-LSTM achieves comparable performance with S-Res-LSTM. It is mainly because S-Res-LSTM achieves the modality adaptation in the finest granularity. The distance between different categories and domains are also narrowed after sample-level adaptation. Thus, the training with jointly domain-, category-, and sample-level adaptation schemes does not bring more improvement. For the MSR dataset, we adopt different levels of adaptation schemes. Compared with baseline Res-LSTM, S-Res-LSTM improves the results on RGB and optical flow by 6.9\% and 5.6\%, respectively.
}

\begin{figure}[b]
\begin{center}

\begin{subfigure}[t]{1.0\linewidth}
		\centering\includegraphics[width=1.0\linewidth]{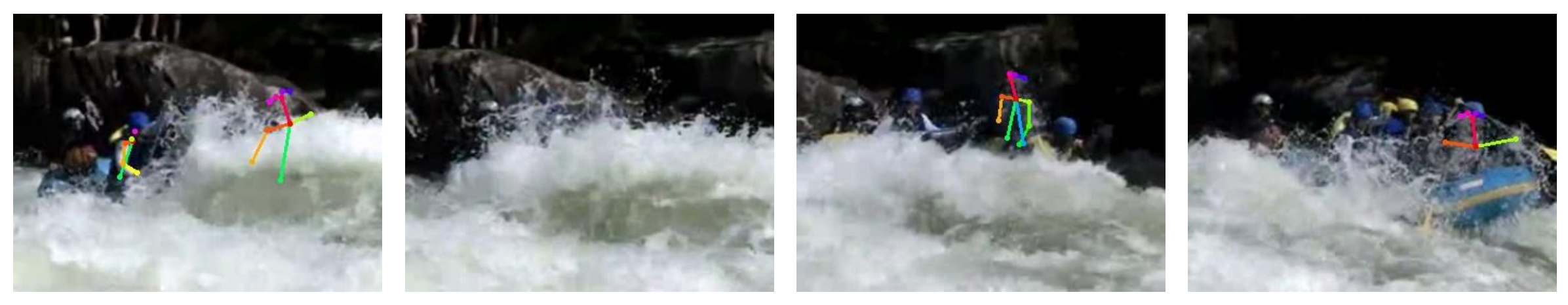}
		\caption{Rafting}			
\end{subfigure}

\begin{subfigure}[t]{1.0\linewidth}
		\centering\includegraphics[width=1.0\linewidth]{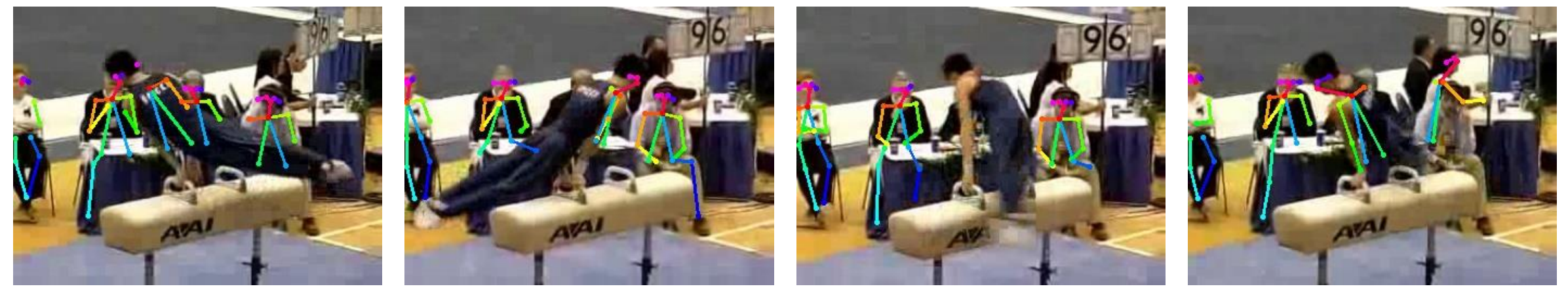}
		\caption{Pommel Horse}			
\end{subfigure}
\caption{Examples from the UCF-101 dataset: (a) rafting, (b) pommel horse. It is challenging to extract high-quality poses with~\cite{cao2017realtime} due to cluttered backgrounds and occlusions.}
\label{fig:ucf_fail_pose}
\end{center}
\end{figure}

\R{Table~\ref{table:result_adap_ucf_jhmdb} shows the results for the UCF-101 and JHMDB datasets with different auxiliary data sources. Note Res-LSTM is our baseline without using any auxiliary data. When using 3D skeletons from the NTU (CV) dataset, it can be regarded as the general case since the color videos are not aligned with skeletons. We use the domain-level adaptation on these two datasets (D-Res-LSTM$^*$). Thanks to the introduction of adaptive representation learning guided by additional skeletons, the common spaces between two different modalities are explored. Thus, we are able to take advantage of the complementary information and improve the feature learning even if the two modalities are not aligned. The results of D-Res-LSTM$^*$ are improved about 1\%-2\% compared with the baseline Res-LSTM on both datasets. When using 2D poses extracted with~\cite{cao2017realtime}, it can be regarded as the specific case and we are able to apply different levels of modal adaptation. The aligned poses provide much more specific guidance. It is observed that on the JHMDB dataset, the sample-level adaptation brings 7.0\% and 3.3\% improvement on RGB and optical flow, respectively. However, on the UCF dataset, the cluttered backgrounds and occlusions make it challenging to extract high-quality human poses, as shown in Fig.~\ref{fig:ucf_fail_pose}. It is less satisfactory to use extracted poses as auxiliary data. Besides, similar as Table~\ref{table:result_adap_ntu_msr}, the jointly adaptation scheme does not help boost the performance.}

\begin{table}[t]
	\begin{center}
		\caption{Performance evaluation on the NTU and MSR datasets in accuracy (\%). The superscript $^*$ denotes the auxiliary data are from external sources. For the NTU dataset, the external source is the PKU dataset.}
        \label{table:result_adap_ntu_msr}
		\begin{tabular}{c|c|c|c|c|c|c}
			\hline
			\multirow{2}{*}{Methods} &\multicolumn{2}{c|}{NTU-CS} &\multicolumn{2}{c|}{NTU-CV} & \multicolumn{2}{c}{MSR}  \\
            \cline{2-7}
             & RGB & Flow & RGB & Flow & RGB & Flow\\
			\hline
            \hline
			 Res-LSTM & 79.6 & 85.8 & 85.7 & 91.8& 67.5 & 71.3\\
            \hline
			 D-Res-LSTM$^*$ & 80.9 & 86.6 & 85.1 & 91.9 & -- & --\\
			\hline
			 C-Res-LSTM$^*$ & 81.4 & 86.8 & 85.2 & 92.1 & -- & --\\
			\hline
			 D-Res-LSTM & 81.4 & 86.4 & 87.3 & 92.0 & 72.5 & 72.5\\
			\hline
			 C-Res-LSTM & 81.5 & 86.6 & 87.2 & 92.1 & 71.9 & 75.0\\
			\hline
			 S-Res-LSTM & \textbf{82.0} & \textbf{87.6} & \textbf{89.1} & 93.3 & \textbf{74.4} & \textbf{76.9}\\
			\hline
             J-Res-LSTM & 81.8 & 87.3 & \textbf{89.1} & \textbf{93.8} & 73.8 & 75.0\\
			\hline
		\end{tabular}
	\end{center}	
\end{table}

\begin{table}[t]
	\begin{center}
		\caption{\R{Performance evaluation with different auxiliary data sources in training on the UCF-101 (split1) and JHMDB (split1) datasets in accuracy (\%). The superscript $^*$ denotes the auxiliary data are from external sources (\emph{i.e.}, the NTU (CV) dataset).}}
\vspace{-2mm}
		\label{table:result_adap_ucf_jhmdb}
		\begin{tabular}{c|c|c|c|c}
			\hline
			\multirow{2}{*}{Methods}  & \multicolumn{2}{c|}{UCF-101} &\multicolumn{2}{c}{JHMDB}   \\
            \cline{2-5}
              & RGB & Flow & RGB & Flow \\
			\hline
            \hline
			 Res-LSTM  &81.6 & 83.8 & 50.9 & 61.5\\
			\hline
			 D-Res-LSTM$^*$  &\textbf{83.3} & \textbf{85.7} & 53.2 & 62.6\\
			\hline
			 D-Res-LSTM  & 79.5 & 84.0 & 52.7 & 62.9 \\
            \hline
             C-Res-LSTM  & 78.4 & 84.7 & 54.9 & 63.6\\
            \hline
             S-Res-LSTM  & 81.9 & 85.3 & \textbf{57.9} & \textbf{64.8}\\
            \hline
             J-Res-LSTM  & 80.1 & 83.3 & 52.7 & 63.6\\
			\hline
		\end{tabular}
	\end{center}	
\vspace{-6mm}
\end{table}

\subsection{Effectiveness of Residual Subnetwork}
\begin{figure*}[t]
\begin{center}

\begin{subfigure}[t]{0.24\linewidth}
		\centering\includegraphics[width=1.0\linewidth]{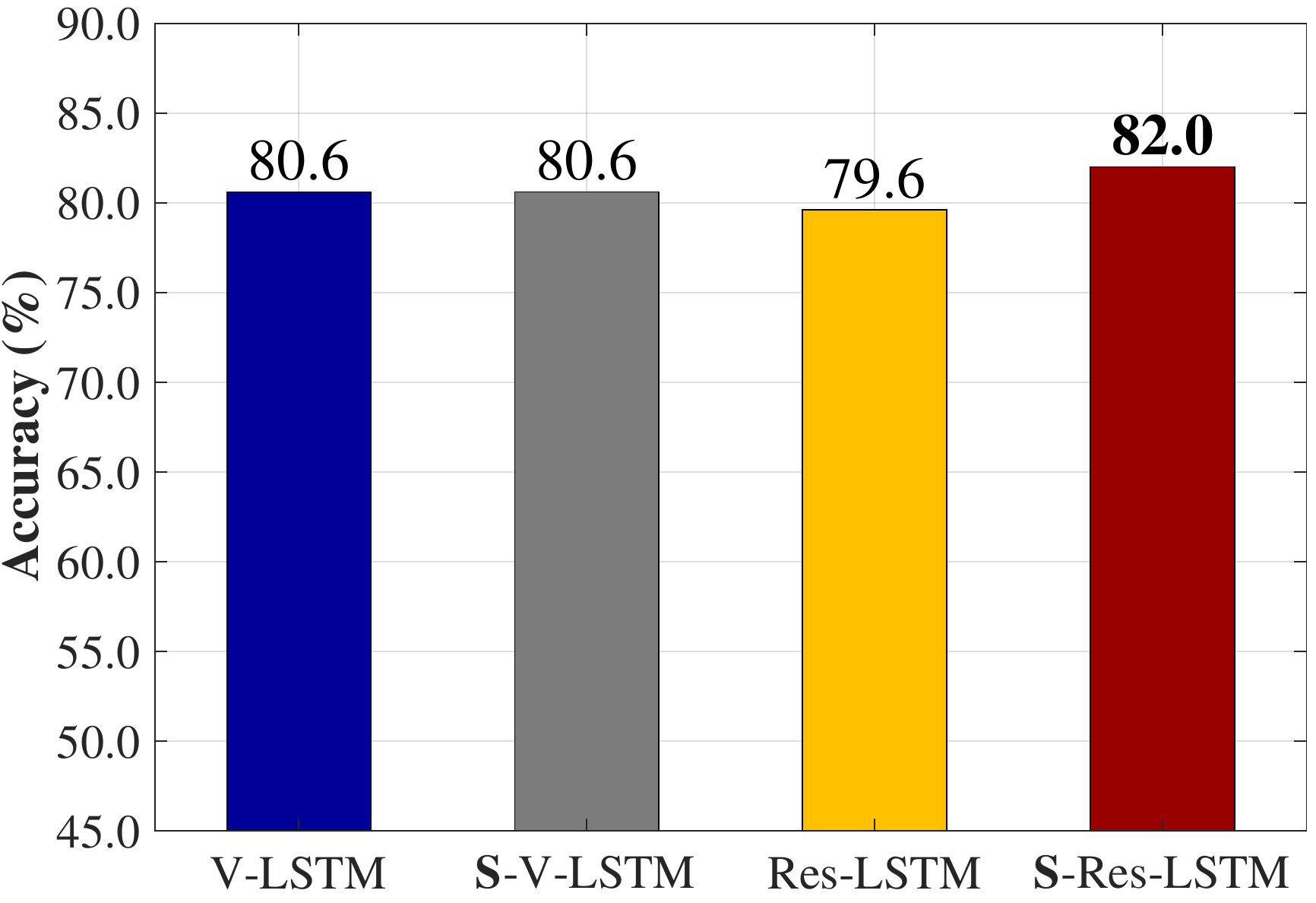}
		\caption{NTU RGB}			
\end{subfigure}
\begin{subfigure}[t]{0.24\linewidth}
		\centering\includegraphics[width=1.0\linewidth]{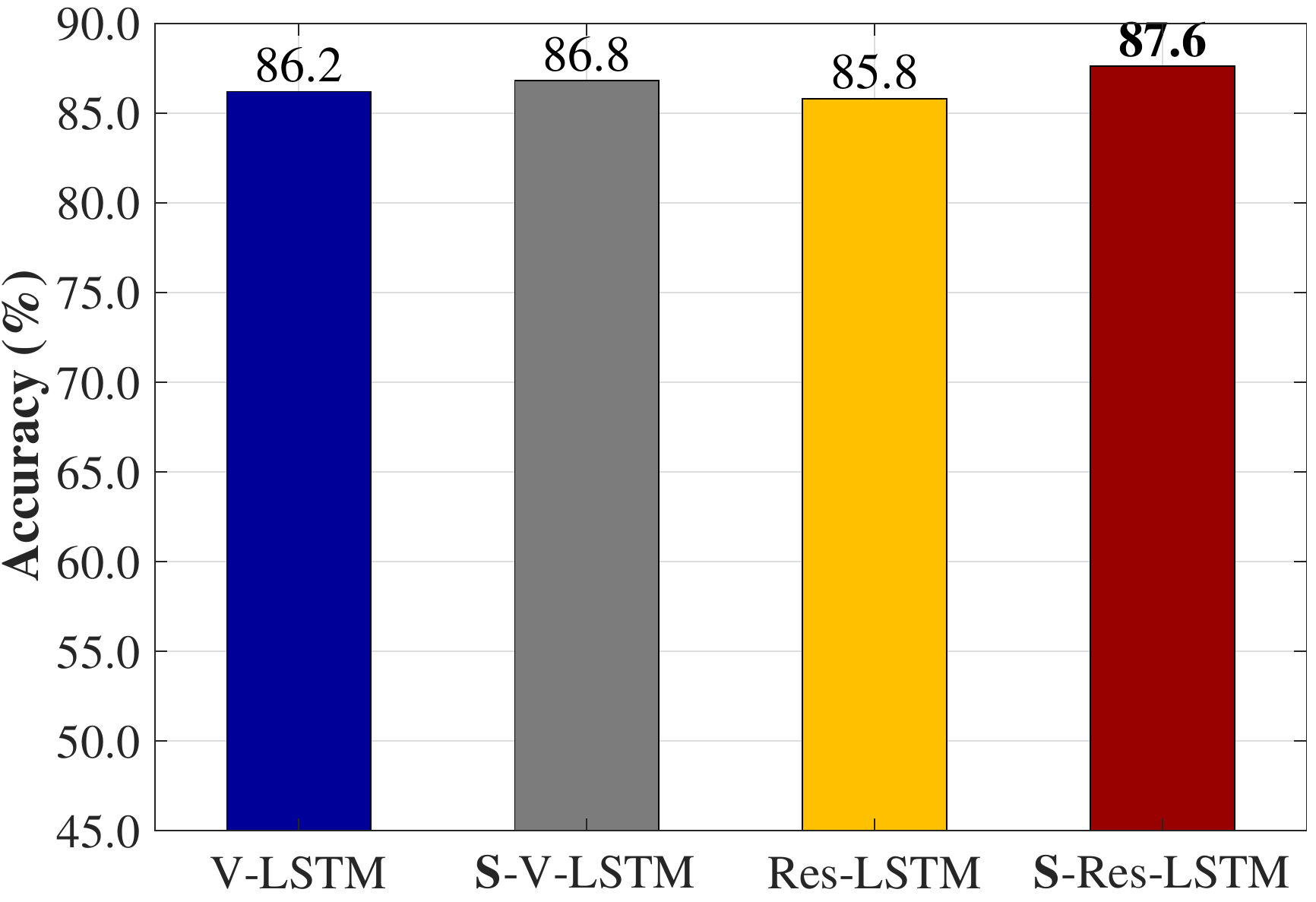}
		\caption{NTU Flow}			
\end{subfigure}
\begin{subfigure}[t]{0.24\linewidth}
		\centering\includegraphics[width=1.0\linewidth]{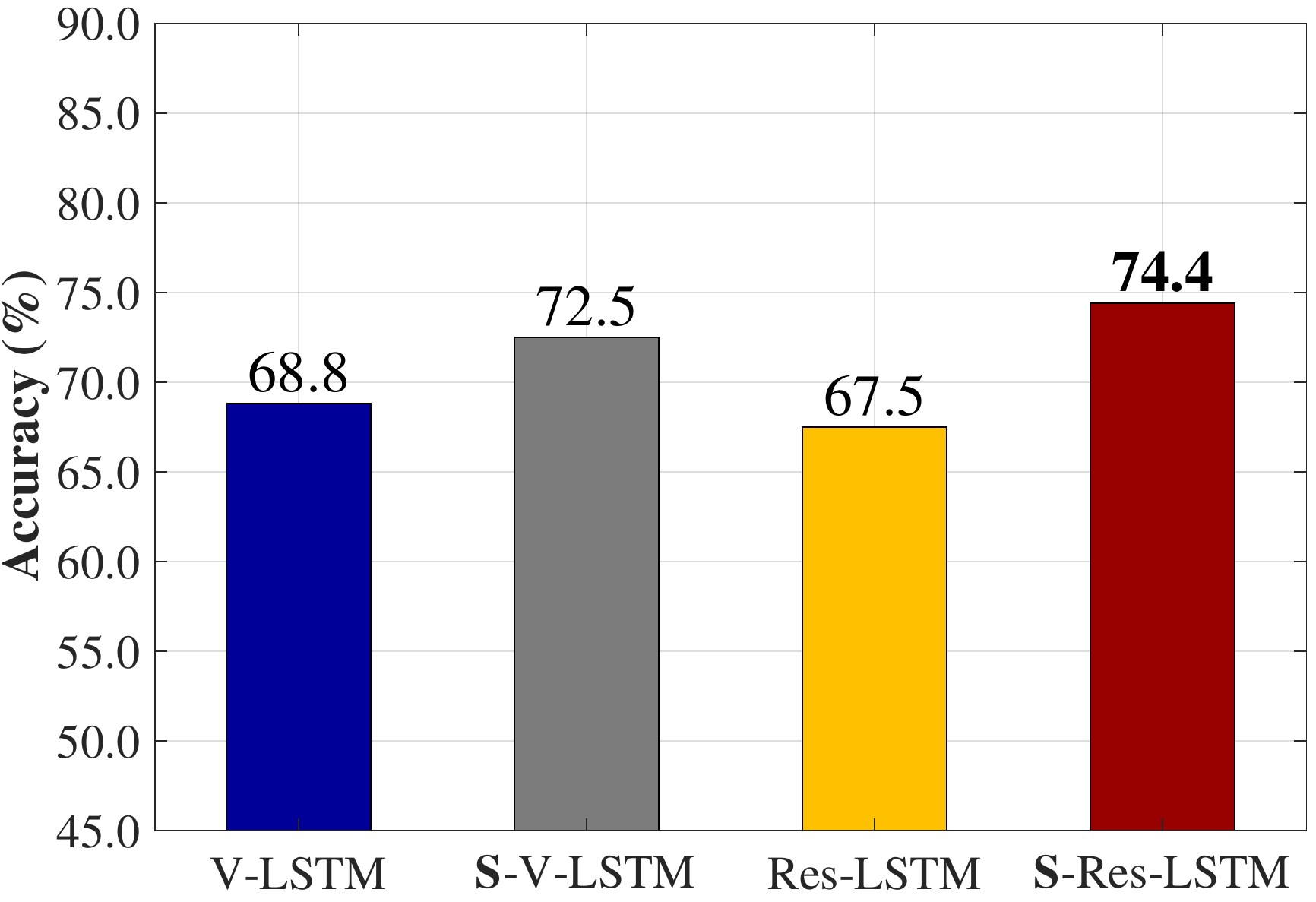}
		\caption{MSR RGB}			
\end{subfigure}
\begin{subfigure}[t]{0.24\linewidth}
		\centering\includegraphics[width=1.0\linewidth]{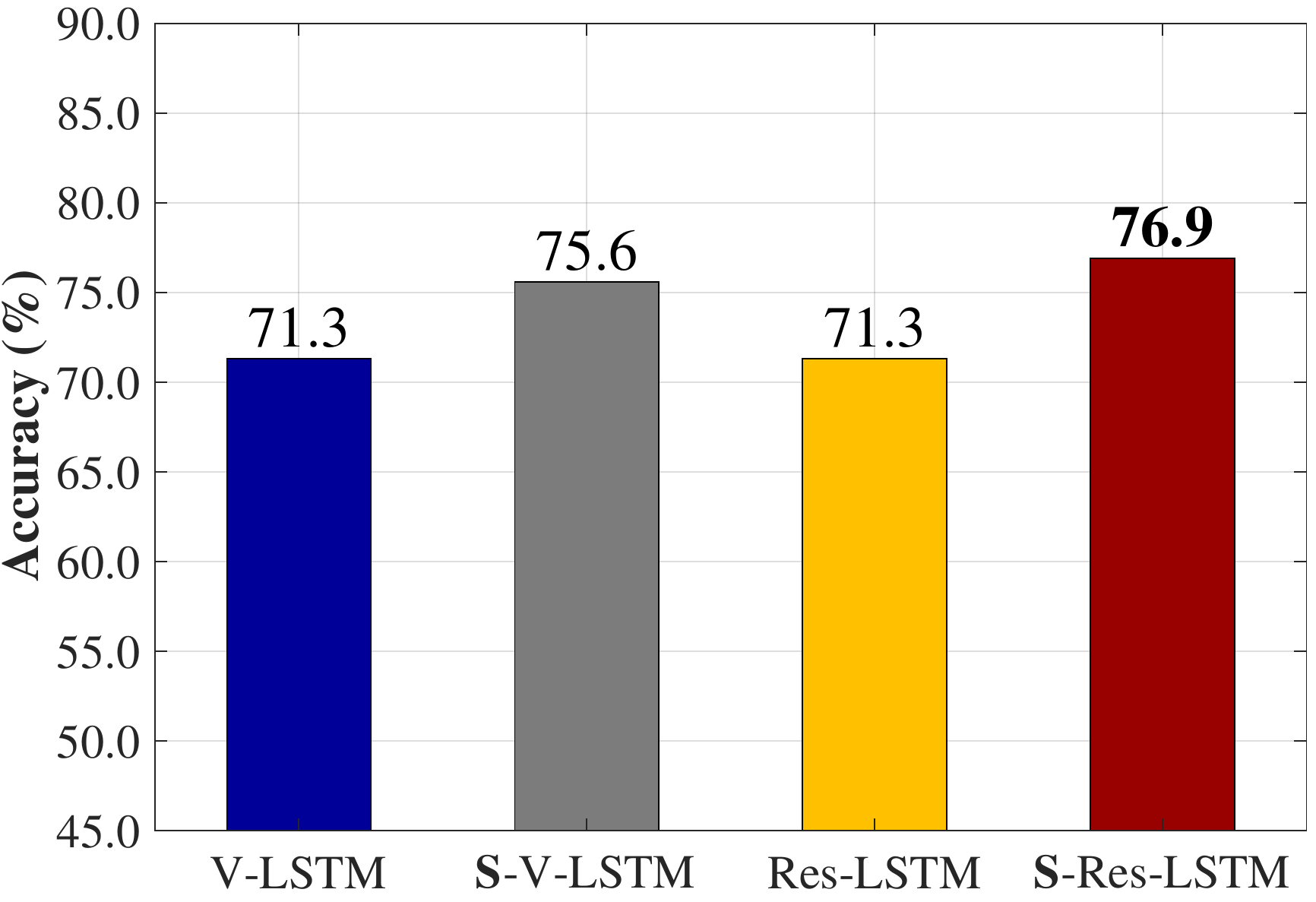}
		\caption{MSR Flow}			
\end{subfigure}

\begin{subfigure}[t]{0.24\linewidth}
		\centering\includegraphics[width=1.0\linewidth]{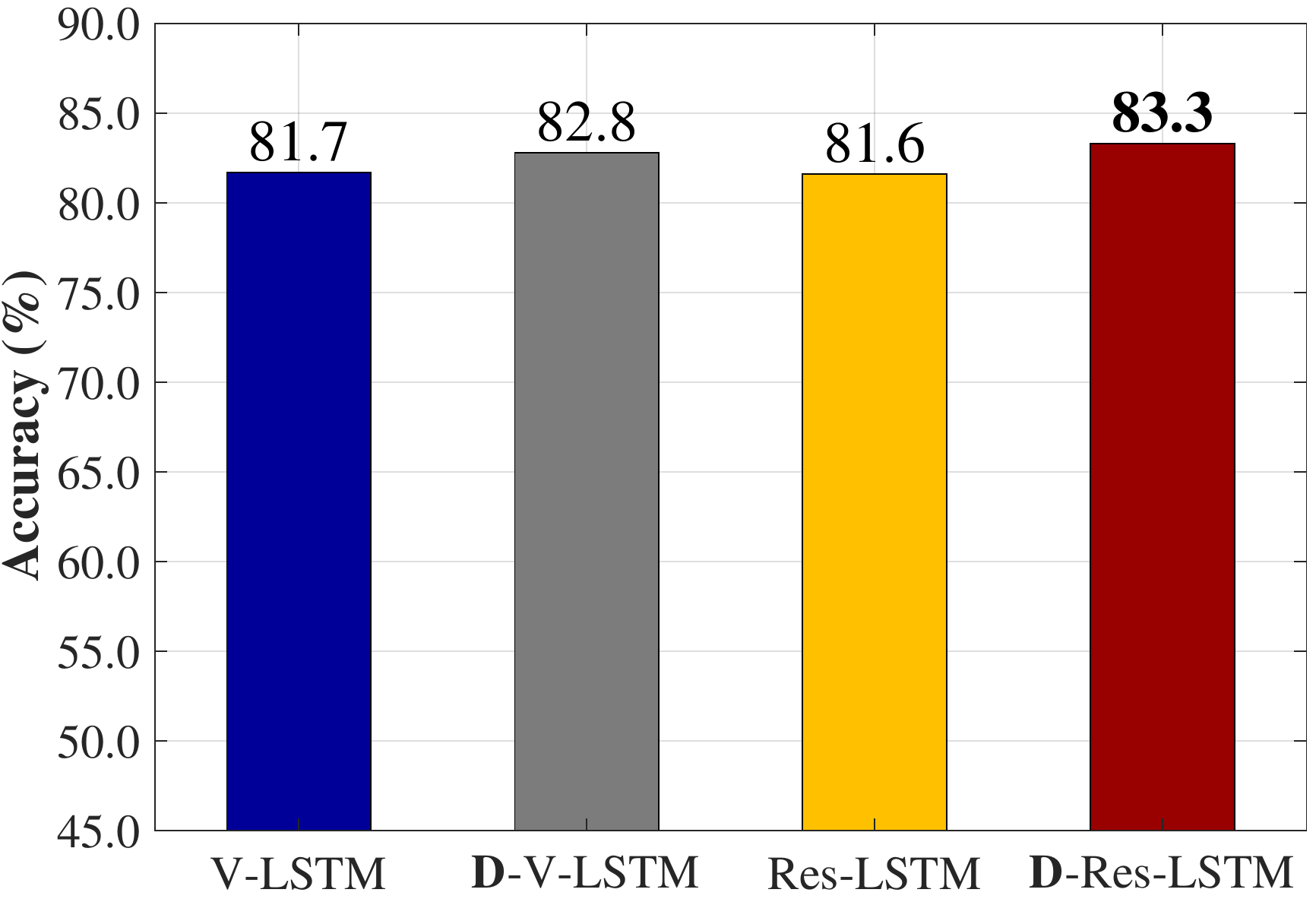}
		\caption{UCF-101 RGB}			
\end{subfigure}
\begin{subfigure}[t]{0.24\linewidth}
		\centering\includegraphics[width=1.0\linewidth]{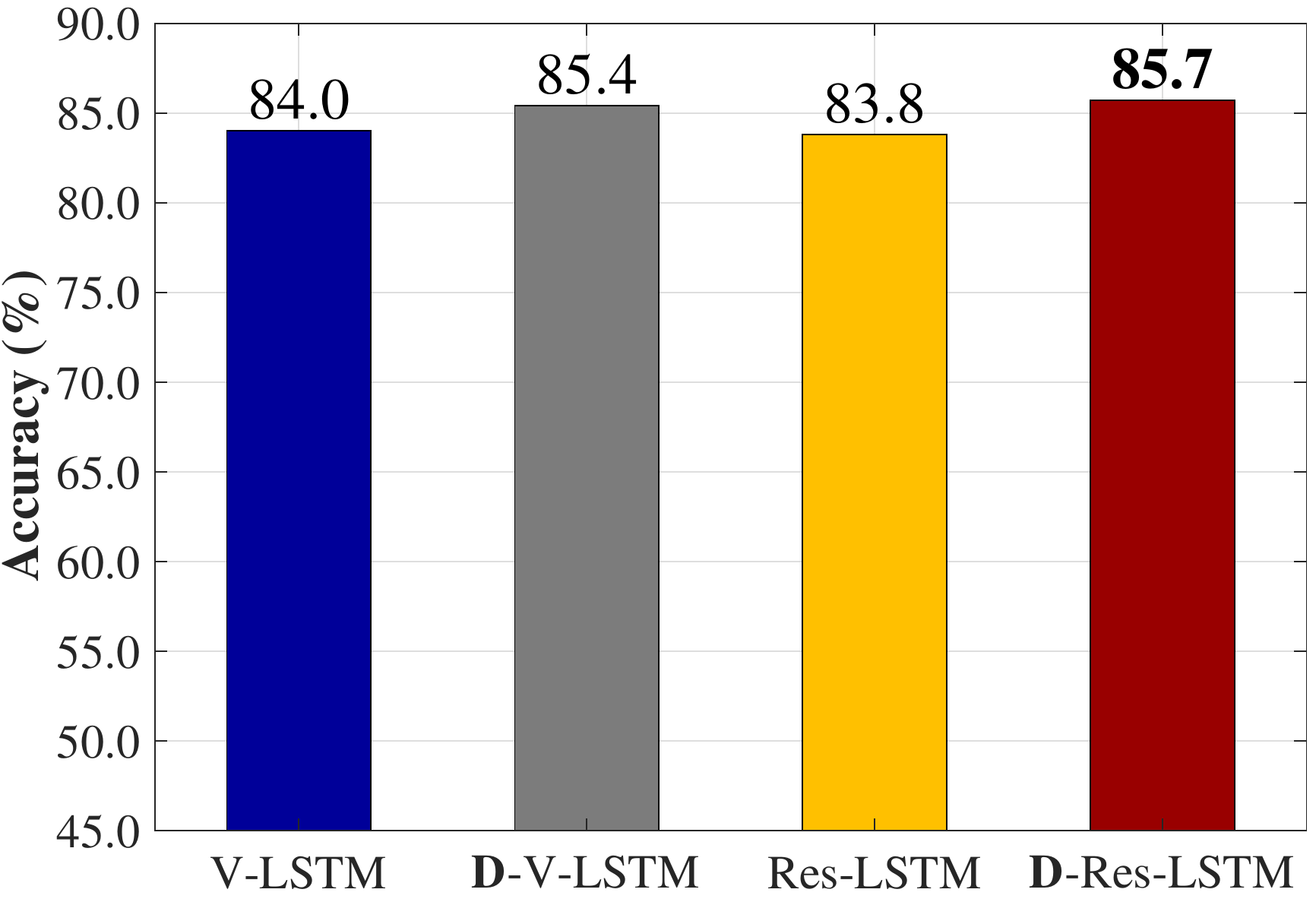}
		\caption{UCF-101 Flow}			
\end{subfigure}
\begin{subfigure}[t]{0.24\linewidth}
		\centering\includegraphics[width=1.0\linewidth]{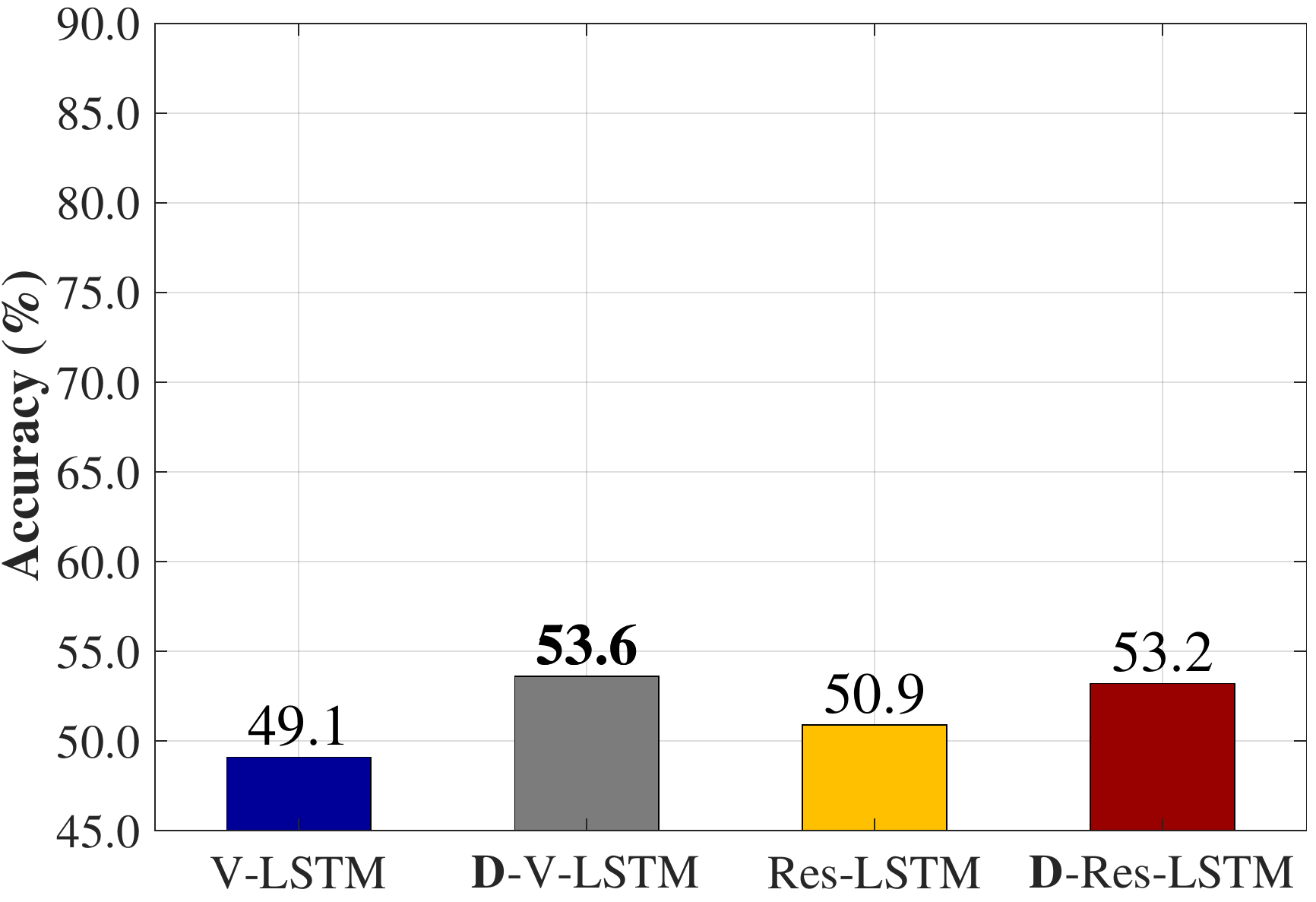}
		\caption{JHMDB RGB}			
\end{subfigure}
\begin{subfigure}[t]{0.24\linewidth}
		\centering\includegraphics[width=1.0\linewidth]{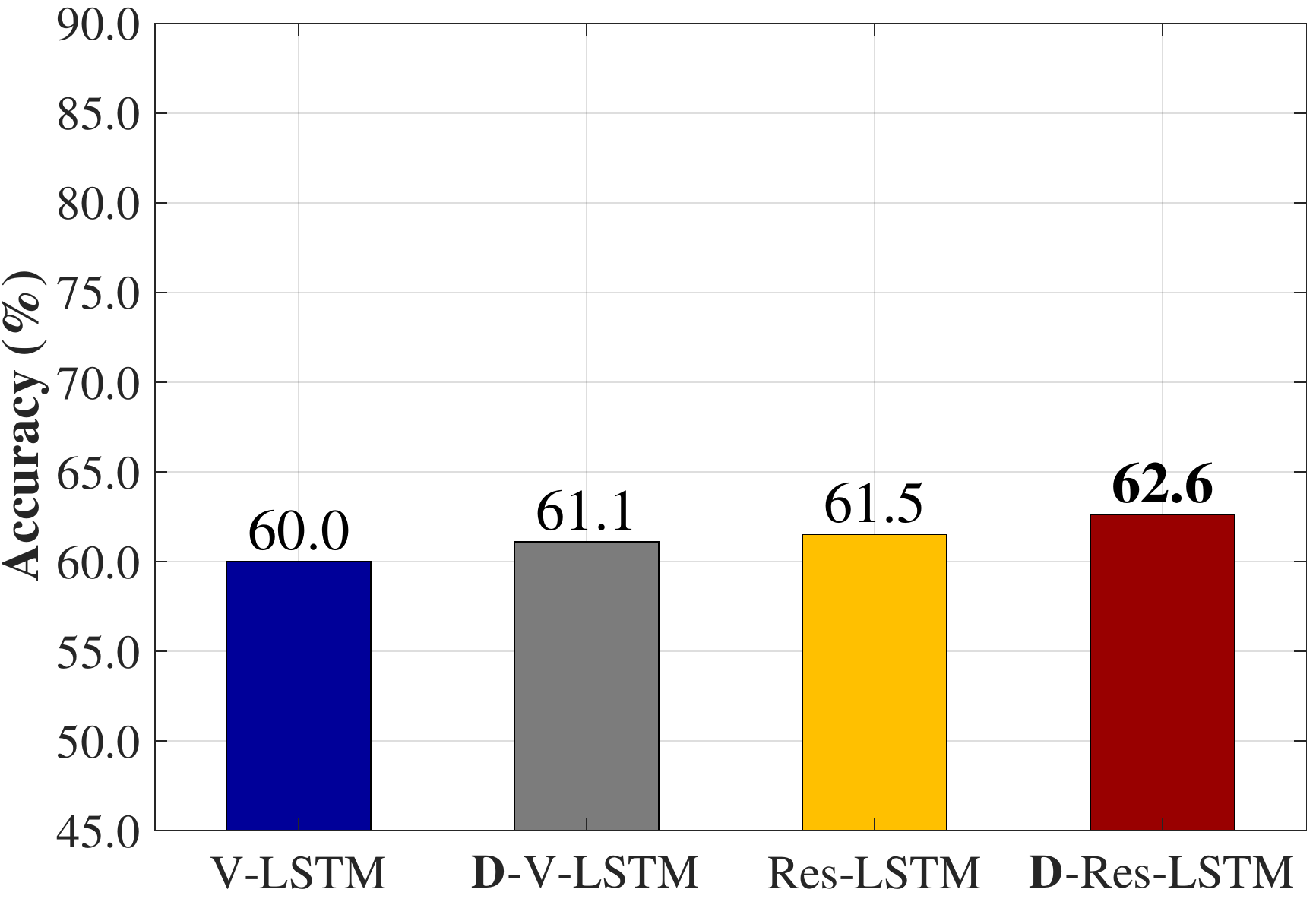}
		\caption{JHMDB Flow}			
\end{subfigure}

\end{center}
\vspace{-2mm}
\caption{\R{Performance comparisons with different structures on different datasets. We apply sample-level adaptation on the NTU (CS) and MSR datasets, domain-level adaptation on the UCF-101 (split1) and JHMDB (split1) datasets, respectively. Our results show that modal adaptation can effectively improve the performance on Res-LSTM and V-LSTM. And the residual architecture is more efficient in compensating the feature learning and achieves better results. }}
\vspace{-2mm}
\label{fig:struct_comp}
\end{figure*}

The design of our residual subnetwork in Res-LSTM is motivated by two aspects: (1) The skip connection in the residual block aims to keep the original information from the source modalities. (2) The residual path is appended to incorporate compensated information, which is adapted from the auxiliary modality. To explore the effectiveness of the residual subnetwork and support our motivation, we compare Res-LSTM with another vanilla LSTM structure V-LSTM, as shown Fig.~\ref{fig:baseline}(b). There are $N$ units for the LSTM layers and the following FC layer, so the number of parameters in V-LSTM is the same as Res-LSTM.

We conduct experiments on these two structures on the four datasets, as shown in Fig.~\ref{fig:struct_comp}. For V-LSTM, the modality adaptation is applied after the second LSTM layer. We adopt sample-level adaptation (S-V-LSTM) for the NTU and MSR datasets, and the domain-level scheme (D-V-LSTM) for the UCF-101 and JHMDB datasets. We obtain similar performance on V-LSTM and Res-LSTM when not using modality adaptation. However, once integrated with modality adaptation, better results are obtained with the architecture of Res-LSTM. The results illustrate that the residual subnetwork is more powerful and efficient in compensating auxiliary information for source data. Besides, our proposed modality adaptation is able to improve the performance with V-LSTM in most cases, which further confirms the effectiveness of the adaptive representation learning.

\subsection{Comparisons to Other State-of-the-Arts}

\begin{table}[b]
\fontsize{7.5pt}{8pt}\selectfont\centering
	\begin{center}
		\caption{Comparisons on NTU in accuracy (\%).}
		\label{table:comp_ntu}
        \begin{tabular}{c|c|c|c}
          \hline
          Methods  & Modality & CS & CV\\
          \hline
          \hline
          Trust Gate~\cite{liu2016spatio} & Ske.  & 69.2 & 77.7\\
          \hline
          STA-LSTM~\cite{song2016end}&Ske.& 73.4 &  81.2 \\
          \hline
          VA-LSTM~\cite{zhang2017view}&Ske.& 79.4  & 87.6\\
          \hline
          LSTM &Poses &  65.0 & 75.4\\
          \hline
          LSTM & Ske. &  70.4 &  83.4\\
          \hline
          P-CNN~\cite{cheron2015p}& RGB+Flow  & 53.8 & 61.7\\
          \hline
          TSN~\cite{wang2016temporal} & RGB+Flow  & 88.5 & 90.4\\
          \hline
          Chained MT~\cite{zolfaghari2017chained} & Poses+RGB+Flow& 80.8 & --\\
          \hline
          \hline
          Res-LSTM &  RGB+Flow & 88.8 & 94.1\\ 
          \hline
          Compensate-Train-Test Res-LSTM & Poses+RGB+Flow & 88.7 & 94.4  \\
          \hline
          Compensate-Train-Test Res-LSTM & Ske.+RGB+Flow & 89.3 & 94.9  \\
          \hline
          S-Res-LSTM &  RGB+Flow & \textbf{89.5} & \textbf{95.2}\\ 
          \hline
          Compensate-Train-Test S-Res-LSTM& Poses+RGB+Flow & 89.4  & 95.4\\
          \hline
          Compensate-Train-Test S-Res-LSTM& Ske.+RGB+Flow& \textbf{90.0}  & \textbf{96.3}\\ 
          \hline
        \end{tabular}
	\end{center}
\vspace{-2mm}
\end{table}

\begin{table}[htbp]
	\begin{center}
		\caption{\M{Comparisons on MSR in accuracy (\%).}}
		\label{table:comp_msr}
        \begin{tabular}{c|c|c}
          \hline
          Methods  & Modality & Acc.(\%) \\
          \hline
          \hline
          Action Ensemble~\cite{wang2012mining} & Ske. & 68.0 \\
          \hline
          Moving Poselets~\cite{tao2015moving} & Ske. & 74.5  \\
          \hline
          LSTM & Poses & 63.8\\
          \hline
          LSTM & Ske. & 72.5 \\
          \hline
          P-CNN~\cite{cheron2015p}&  RGB+Flow  & 61.9 \\
          \hline
          TSN~\cite{wang2016temporal} & RGB+Flow   & 88.1\\
          \hline
          \hline
          Res-LSTM &  RGB+Flow & 84.4 \\ 
          \hline
          Compensate-Train-Test Res-LSTM & Poses+RGB+Flow & 85.0 \\ 
          \hline
          Compensate-Train-Test Res-LSTM& Ske.+RGB+Flow & 85.6 \\ 
          \hline
          S-Res-LSTM &  RGB+Flow & \textbf{88.8} \\ 
          \hline
          Compensate-Train-Test S-Res-LSTM & Poses+RGB+Flow & 90.6 \\ 
          \hline
          Compensate-Train-Test S-Res-LSTM & Ske.+RGB+Flow & \textbf{91.9}  \\ 
          \hline
        \end{tabular}
	\end{center}
\vspace{-2mm}
\end{table}

We compare our final results with other state-of-the-art methods, as shown in Table~\ref{table:comp_ntu}, Table~\ref{table:comp_msr}, Table~\ref{table:comp_ucf}, and Table~\ref{table:comp_jhmdb}. For the specific case, we use S-Res-LSTM as our final model. And for the general case, we regard D-Res-LSTM$^*$ as our final model. In our work, the final prediction for a given video is generated by average score fusion of multiple streams~\cite{feichtenhofer2016convolutional}.

\R{To give a fair comparison, we mark the modalities contributed in the testing phase in Table~\ref{table:comp_ntu}, Table~\ref{table:comp_msr}, Table~\ref{table:comp_ucf}, and Table~\ref{table:comp_jhmdb}. The involved modalities include 3D skeletons (ske.), 2D poses extracted with~\cite{cao2017realtime} (poses), RGB videos (RGB) and optical flow data (Flow). We also show our recognition results only with skeleton/pose data using a 1-layer LSTM (LSTM). Our baseline results include Res-LSTM and Compensate-Train-Test Res-LSTM. For Res-LSTM, only source modalities are employed in the testing phase without any modal adaptation. For Compensate-Train-Test Res-LSTM, skeletons/poses are used in the testing phase and average fusion is employed to fuse RGB, optical flow and skeleton/pose streams.}

\R{In Table~\ref{table:comp_ntu} and Table~\ref{table:comp_msr}, our model S-Res-LSTM not only outperforms the baseline Res-LSTM, but also achieves higher performance than that even skeletons/poses are available in testing (Compensate-Train-Test Res-LSTM). It illustrates that our modal adaption scheme effectively compensates for the loss of skeletons at test time. In the meanwhile, we also find that fusing the results of LSTM and S-Res-LSTM, denoted as Compensate-Train-Test S-Res-LSTM, can further improve the performance. Furthermore, both S-Res-LSTM and Compensate-Train-Test S-Res-LSTM compare favorably to the existing methods on the NTU and MSR datasets. Note that we implement P-CNN~\cite{cheron2015p} and TSN~\cite{wang2016temporal} with the code provided by the authors. The result under the cross-view setting on the NTU dataset is not given in~\cite{zolfaghari2017chained}.}

In Table~\ref{table:comp_ucf} and Table~\ref{table:comp_jhmdb}, we report results over three splits on the UCF-101 and JHMDB datasets, respectively. On the UCF-101 dataset, as indicated in Table~\ref{table:result_adap_ucf_jhmdb} and Fig.~\ref{fig:ucf_fail_pose}, the inaccurate poses lead to poor action classification results, only 36.0\% with LSTM. They also fail to improve the feature learning with S-Res-LSTM. However, D-Res-LSTM$^*$, which is the two-stream fusion result with skeletons in NTU-CV as auxiliary data, outperforms Compensate Train-Test Res-LSTM and S-Res-LSTM. On the JHMDB dataset, we obtain similar results with D-Res-LSTM$^*$ and S-Res-LSTM, though they use different auxiliary data sources. And both of them effectively improve the baseline results. Compared with other state-of-the-arts, our model achieves comparable performance on the UCF-101 dataset and outperforms other methods on the JHMDB dataset. To give a fair comparison, we only compare our results using RGB and optical flow at the test time on both datasets. \RR{Note that the action categories in the UCF-101 dataset are very diverse, including \emph{sky diving}, \emph{rafting}, \emph{surfing}, \emph{etc.}, which are very different from those in the NTU dataset, such as \emph{shaking hands}, \emph{drinking}, \emph{eating}, \emph{etc}. We believe that training with skeletons in more relevant categories will bring further improvement on the UCF-101 dataset through the proposed modality adaption scheme.} 

\begin{table}[t]
	\begin{center}
		\caption{Comparisons on UCF-101 in accuracy (\%). The superscript $^*$ denotes auxiliary data are from external sources (\emph{i.e.}, the NTU (CV) dataset).}
		\label{table:comp_ucf}
        \begin{tabular}{c|c|c}
          \hline
          Methods & Modality & Acc.(\%) \\
          \hline
          \hline
          LSTM & Poses & 36.0 \\
          \hline
          LRCN~\cite{donahue2015long} & RGB & 82.9\\
          \hline
          C3D~\cite{tran2015learning}& RGB & 85.2\\
          \hline
          Two-Stream ConvNet~\cite{simonyan2014two} & RGB+Flow &88.0\\
          \hline
          Two-Stream + Fusion~\cite{feichtenhofer2016convolutional} & RGB+Flow  & 92.5 \\
          \hline
          TSN~\cite{wang2016temporal} &  RGB+Flow  & \textbf{94.0} \\
          \hline
          Chained MT~\cite{zolfaghari2017chained} & RGB+Flow  & 91.3\\
          \hline
          \hline
          Res-LSTM  & RGB+Flow  & 91.7\\
          \hline
          Compensate Train-Test Res-LSTM & Poses+RGB+Flow & 91.8 \\
          \hline
          D-Res-LSTM$^*$ & RGB+Flow & 92.3\\ 
          \hline
          S-Res-LSTM & RGB+Flow & 90.7 \\
          \hline
        \end{tabular}
	\end{center}
\vspace{-6mm}
\end{table}

\begin{table}[htbp]
	\begin{center}
		\caption{Comparisons on JHMDB in accuracy (\%). The superscript $^*$ denotes auxiliary data are from external sources (\emph{i.e.}, the NTU (CV) dataset).}
		\label{table:comp_jhmdb}
        \begin{tabular}{c|c|c}
          \hline
          Methods  &  Modality & Acc.(\%) \\
          \hline
          \hline
          LSTM & Poses & 47.8\\
          \hline
          P-CNN~\cite{cheron2015p}& RGB+Flow& 61.1\\
          \hline
          Action Tubes~\cite{gkioxari2015finding}&RGB+Flow& 62.5\\
          \hline
          TS R-CNN~\cite{peng2016multi}& RGB+Flow& 70.5\\
          \hline
          MR-TS R-CNN~\cite{peng2016multi}& RGB+Flow& 71.1\\
          \hline
          TSN~\cite{wang2016temporal} & RGB+Flow& 73.4 \\
          \hline
          Chained MT~\cite{zolfaghari2017chained} &RGB+Flow& 72.8\\
          \hline
          \hline
          Res-LSTM &RGB+Flow&  72.7\\
          \hline
          Compensate Train-Test Res-LSTM &Poses+RGB+Flow& 72.9 \\
          \hline
          D-Res-LSTM$^*$  &RGB+Flow&  74.6 \\ 
          \hline
          S-Res-LSTM &RGB+Flow&  \textbf {74.8} \\
          \hline
        \end{tabular}
	\end{center}
\vspace{-6mm}
\end{table}

\subsection{Discussion}

\label{sec:param}
\begin{figure}[!b]
\begin{center}
\begin{subfigure}[b]{1.0\linewidth}
		\centering\includegraphics[width=0.45\linewidth]{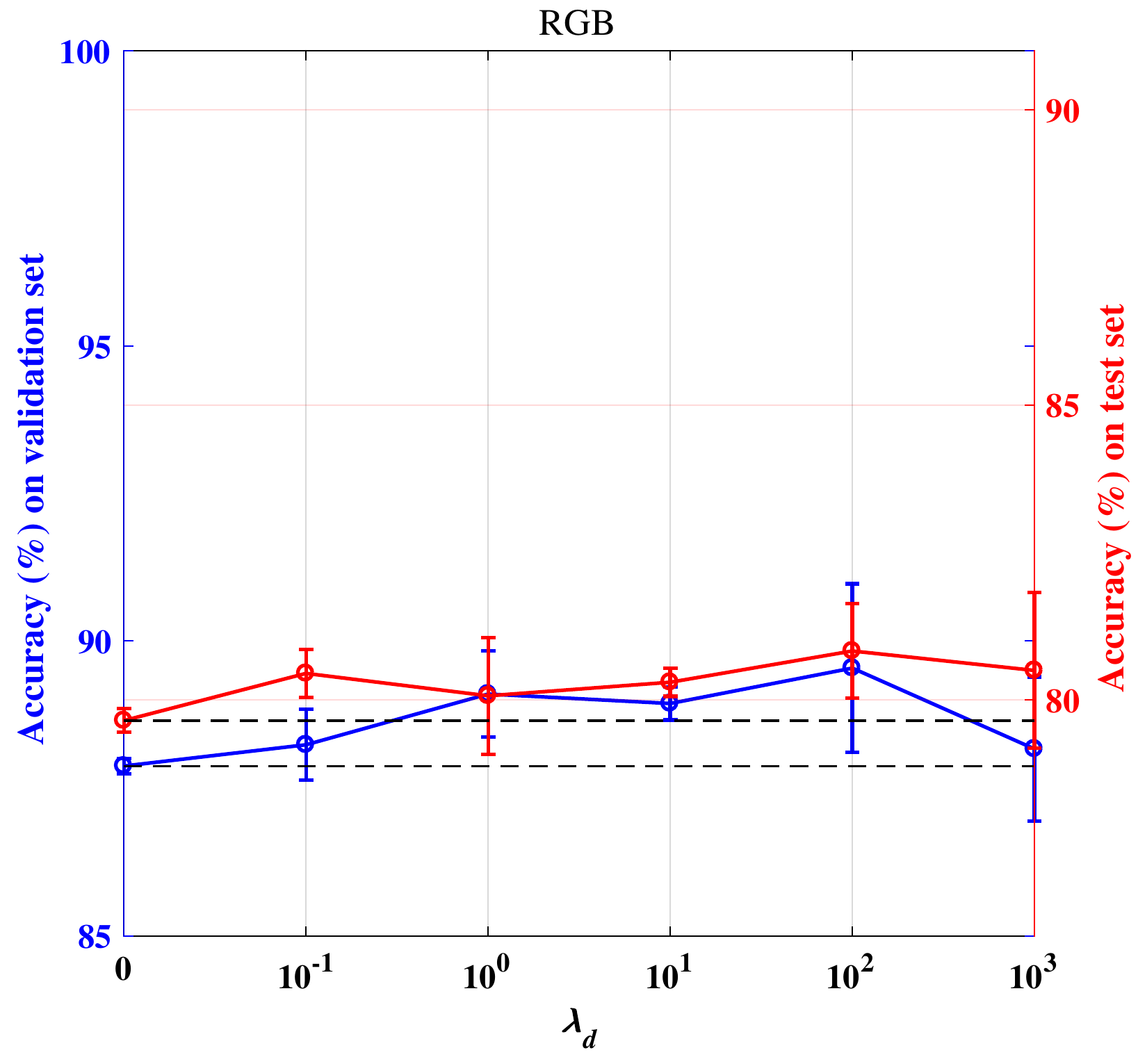}	
        \centering\includegraphics[width=0.45\linewidth]{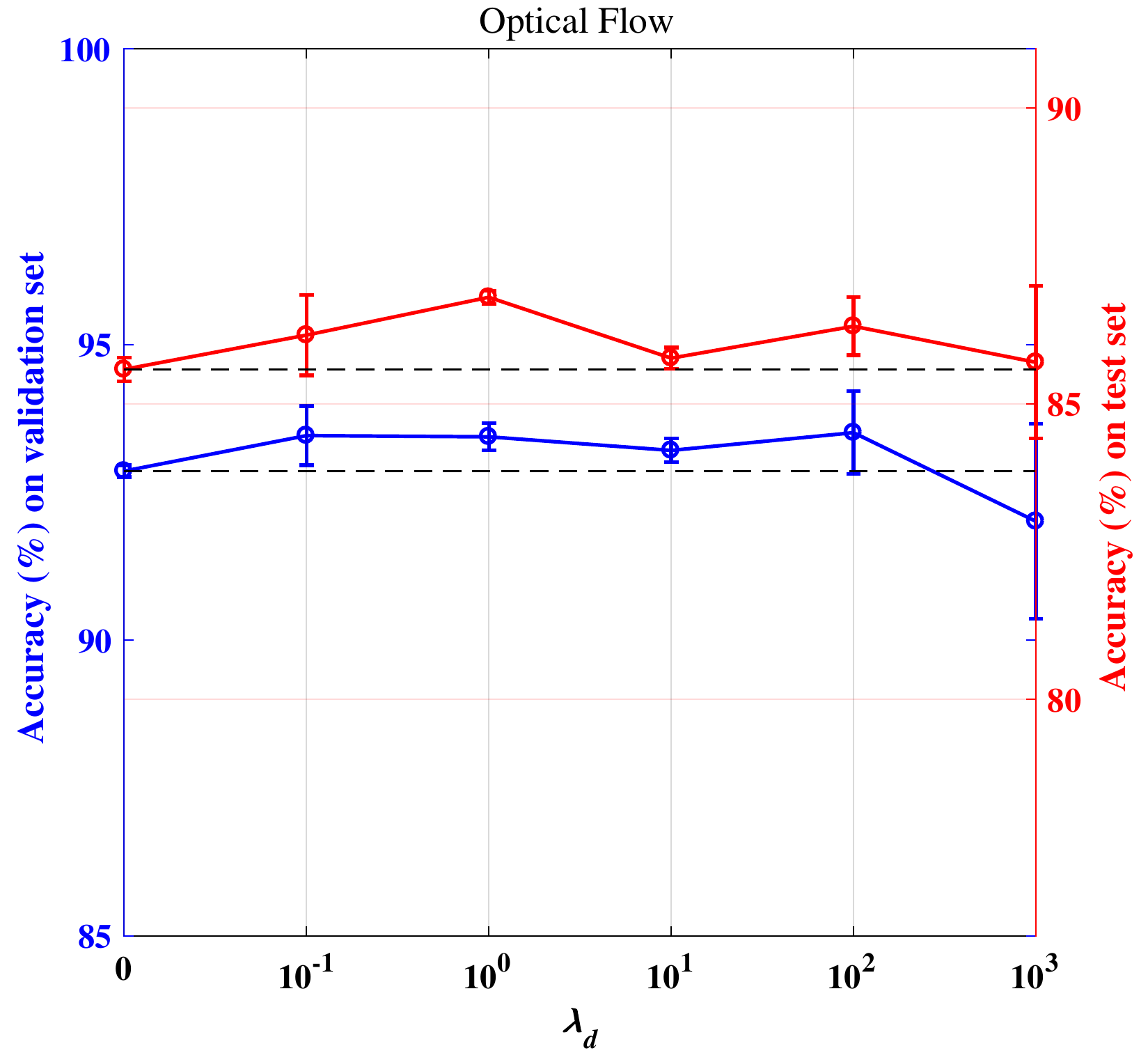}
		\caption{Domain-level adaptation}		
\end{subfigure}

\begin{subfigure}[b]{1.0\linewidth}
		\centering\includegraphics[width=0.45\linewidth]{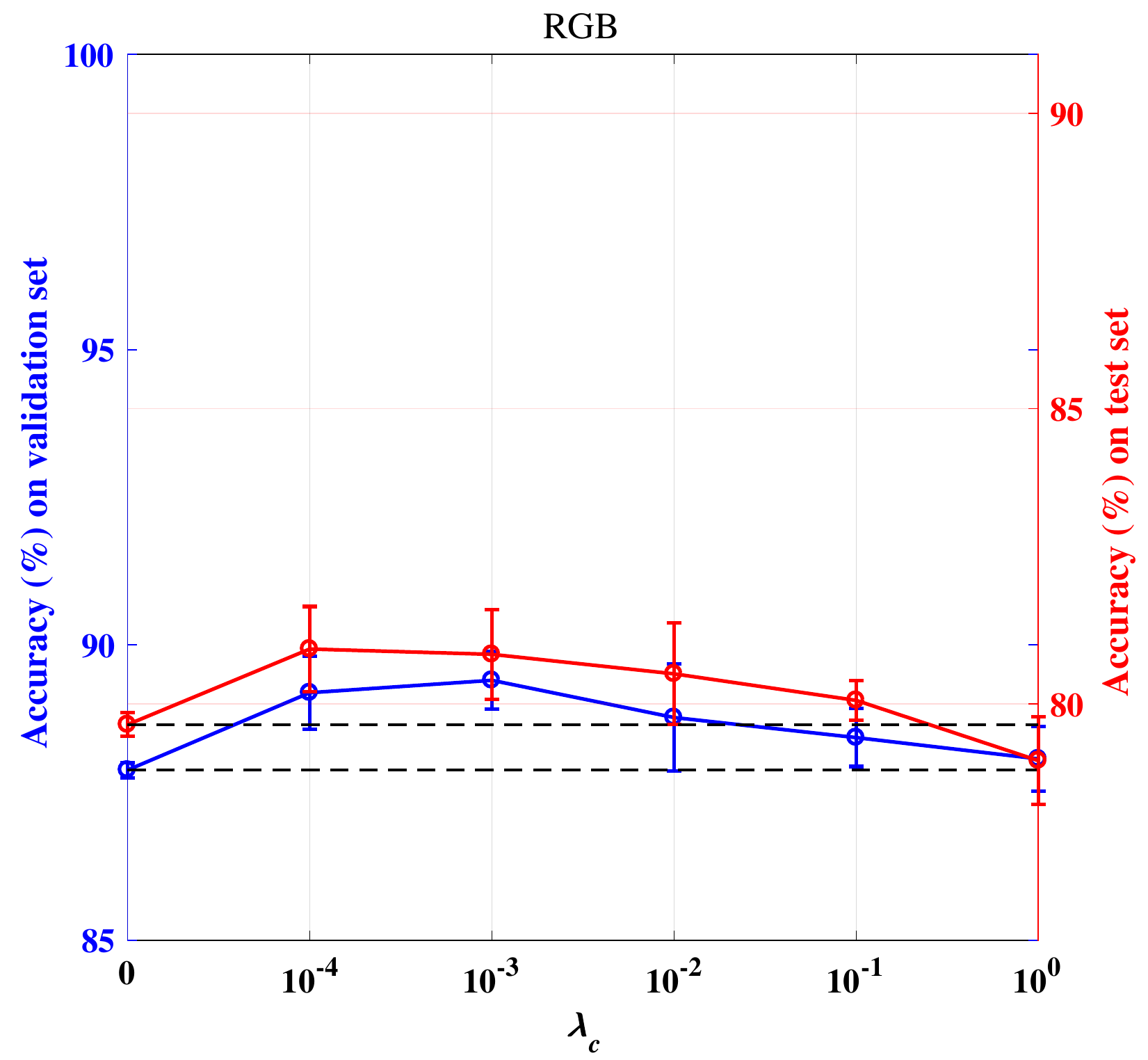}	
        \centering\includegraphics[width=0.45\linewidth]{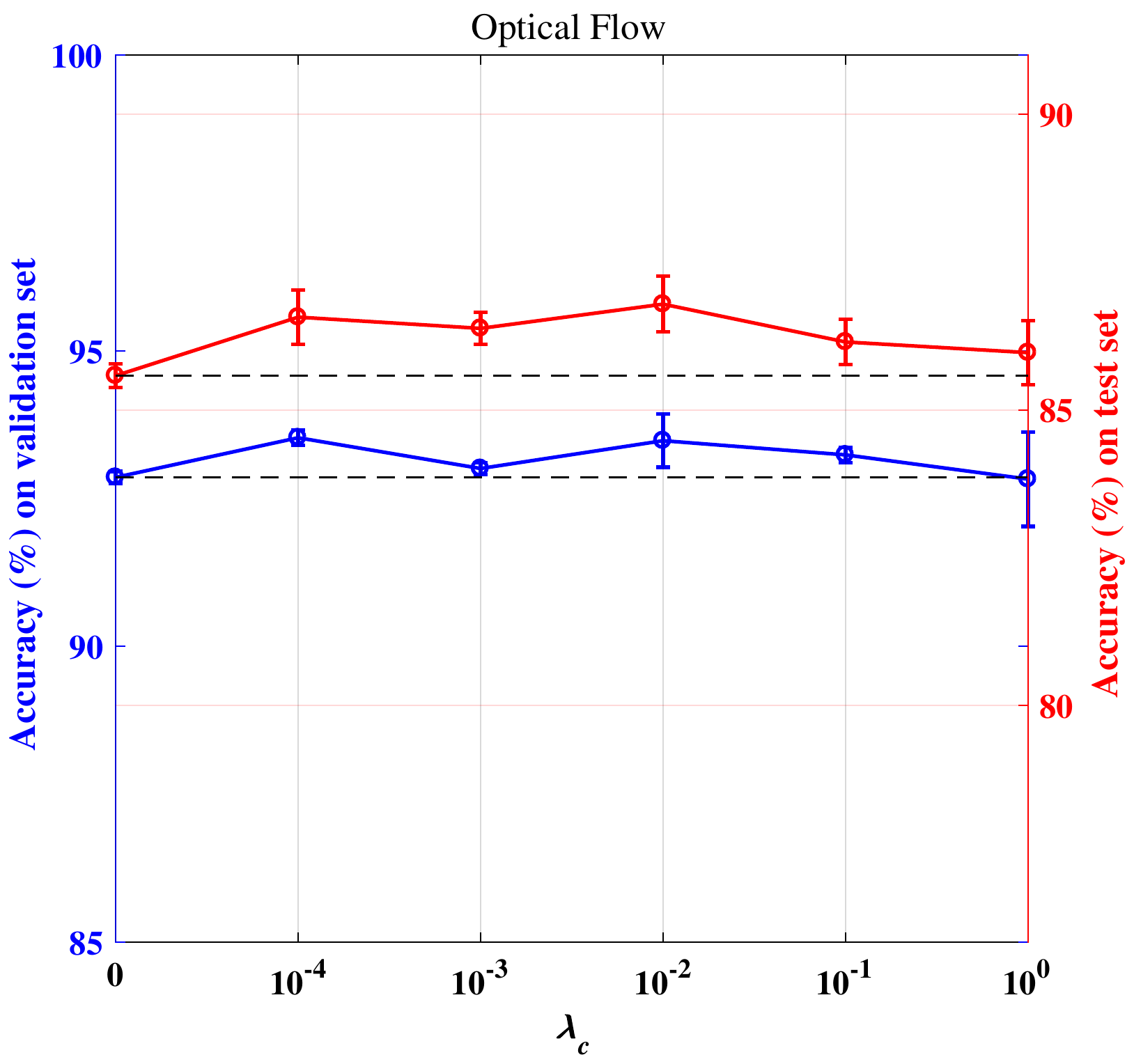}
        \caption{Category-level adaptation}		
\end{subfigure}

\begin{subfigure}[b]{1.0\linewidth}
		\centering\includegraphics[width=0.45\linewidth]{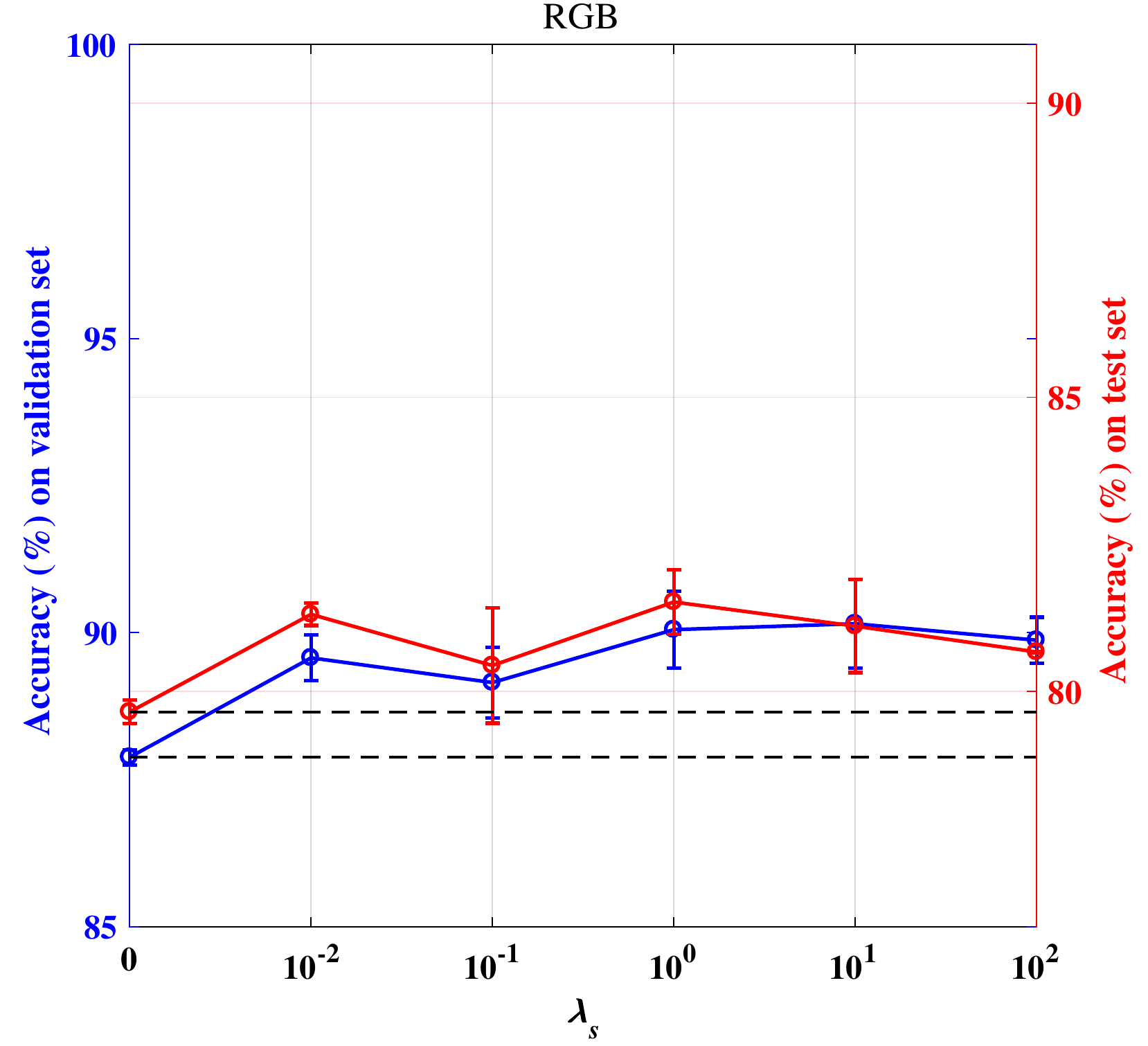}
        \centering\includegraphics[width=0.45\linewidth]{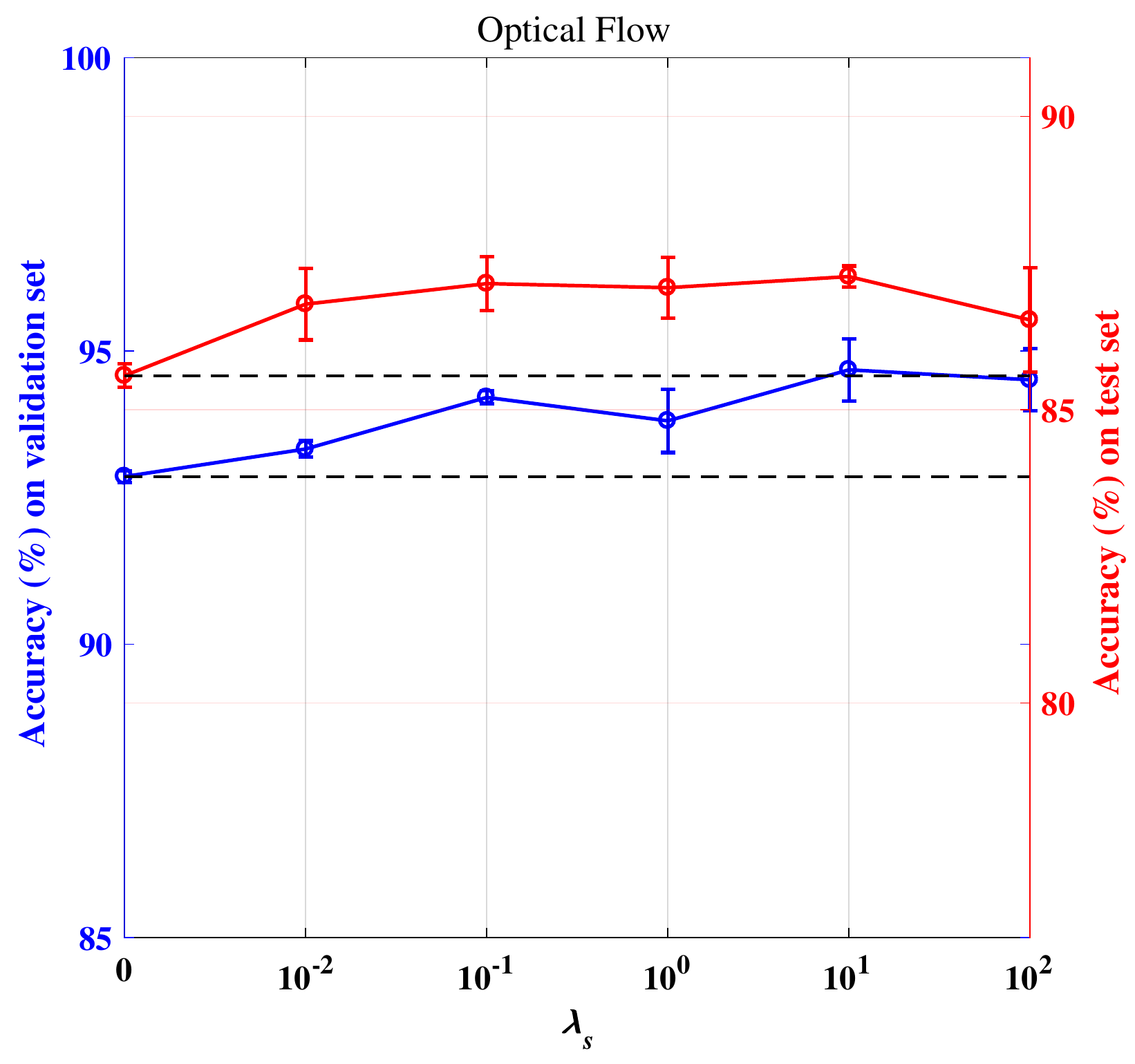}
		\caption{Sample-level adaptation}			
\end{subfigure}

\end{center}
\vspace{-2mm}
\caption{\M{Empirical study on the hyper parameter $\lambda$ of different modal adaptation schemes on the NTU (CS) dataset.}}
\vspace{-1mm}
\label{fig:effect_lambda}
\end{figure}

\begin{figure}[t!]
\begin{center}

\begin{subfigure}[t]{1.0\linewidth}
		\centering\includegraphics[width=1.0\linewidth]{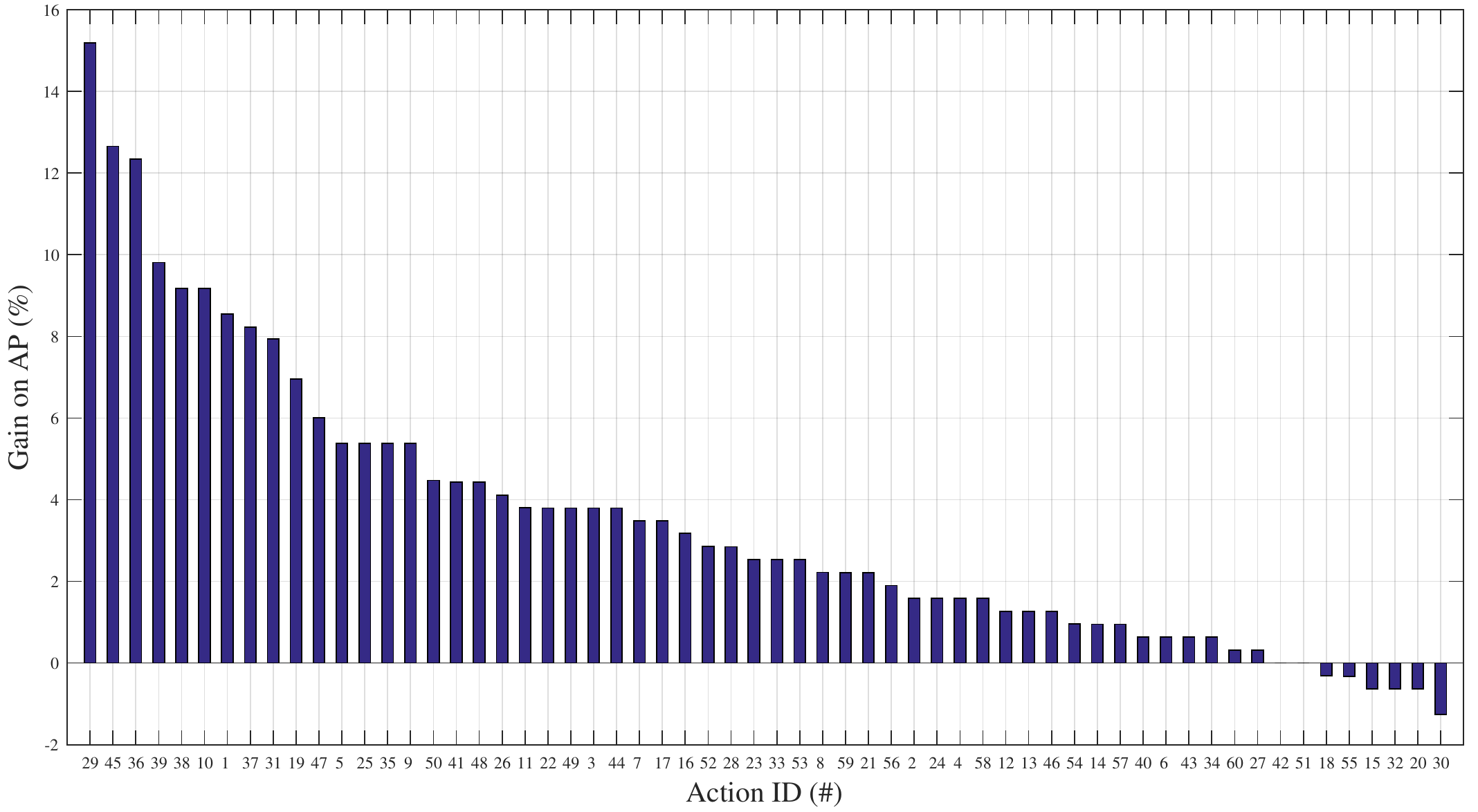}
		\caption{NTU RGB}			
\end{subfigure}

\begin{subfigure}[t]{1.0\linewidth}
		\centering\includegraphics[width=1.0\linewidth]{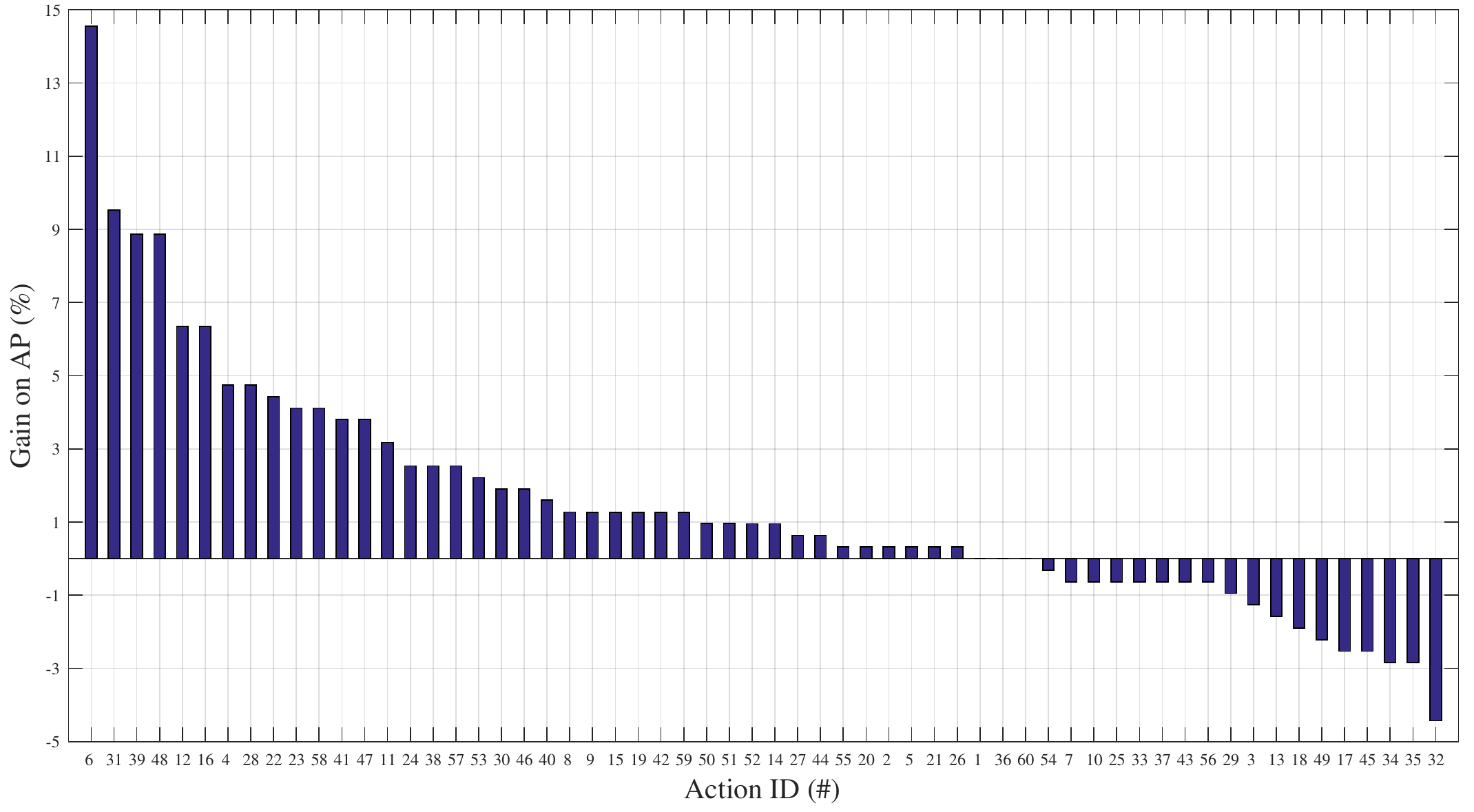}
		\caption{NTU Flow}			
\end{subfigure}

\end{center}
\vspace{-2mm}
\caption{\M{Gain on average precision of S-Res-LSTM with respect to Res-LSTM on the NTU (CV) dataset. The index of the horizontal axis denotes the action ID provided in~\cite{shahroudy2016ntu}.}}
\vspace{-3mm}
\label{fig:classification}
\end{figure}

\begin{figure*}[ht]
\centering
	\begin{subfigure}[t]{0.24\linewidth}
		\centering\includegraphics[width=1.0\textwidth]{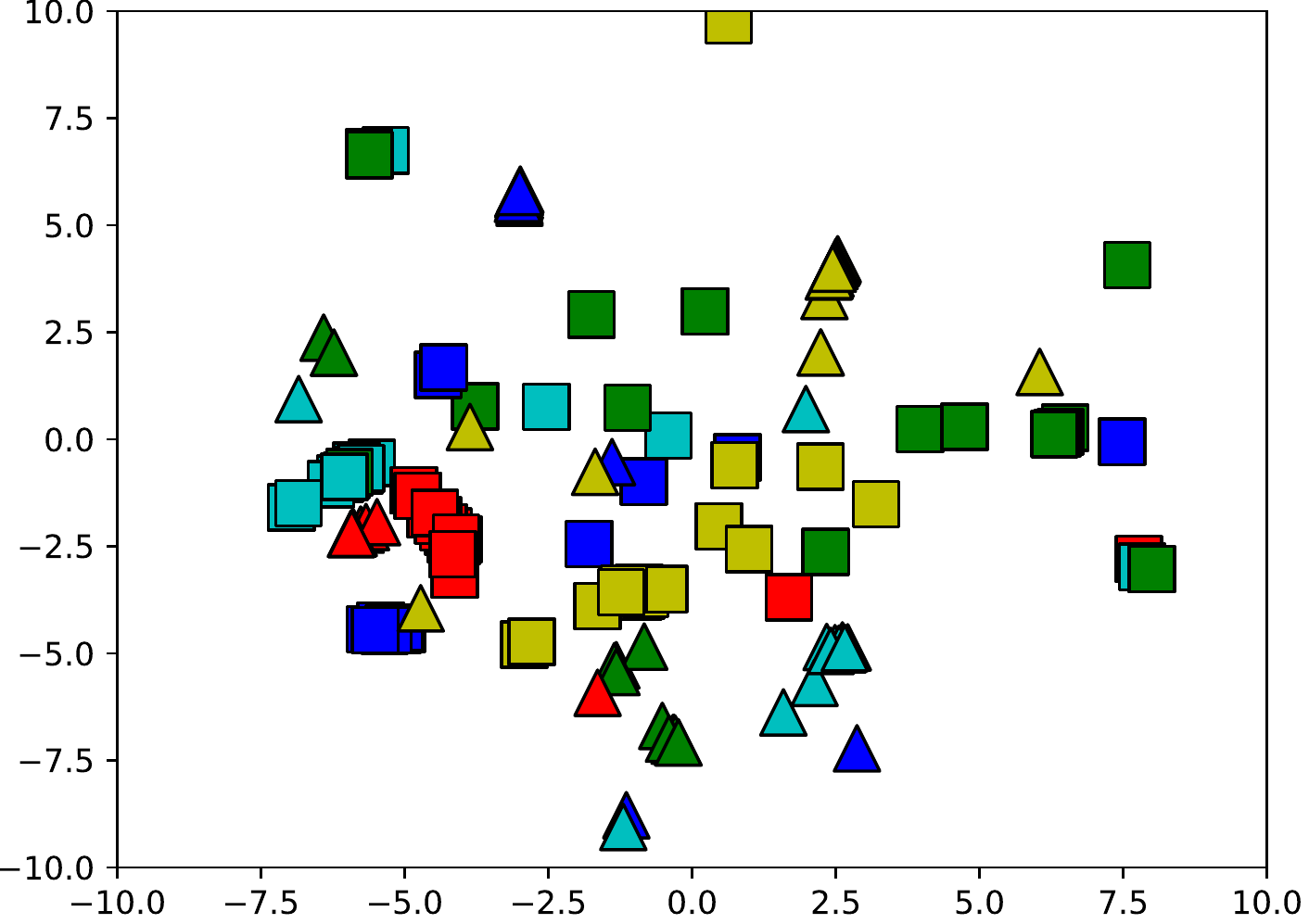}
		\label{fig:baseline_rgb_ske}
	\end{subfigure}
	\begin{subfigure}[t]{0.24\linewidth}
		\centering\includegraphics[width=1.0\textwidth]{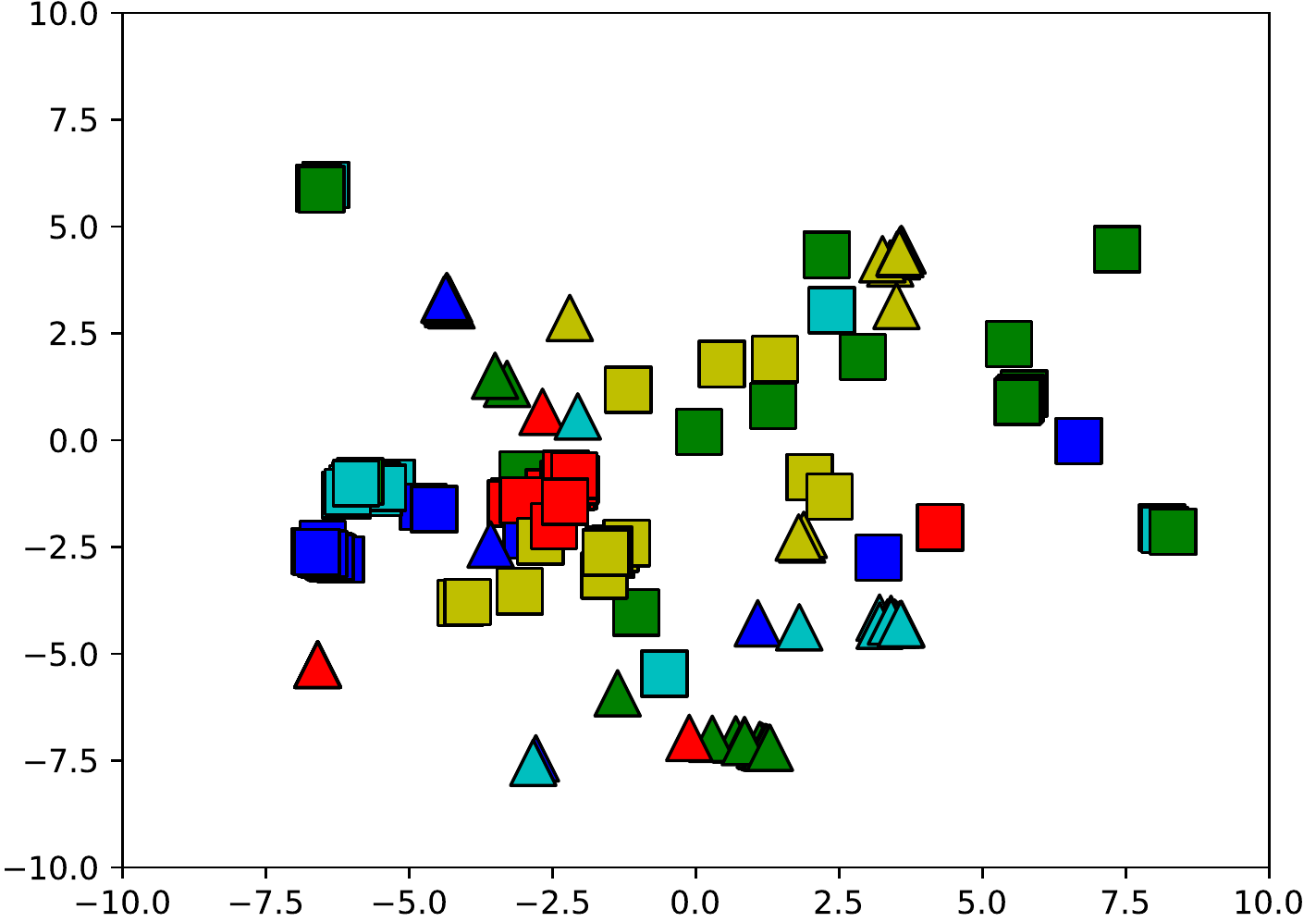}
		
		\label{fig:domain_rgb_ske}
	\end{subfigure}	
    \begin{subfigure}[t]{0.24\linewidth}
    \centering\includegraphics[width=1.0\textwidth]{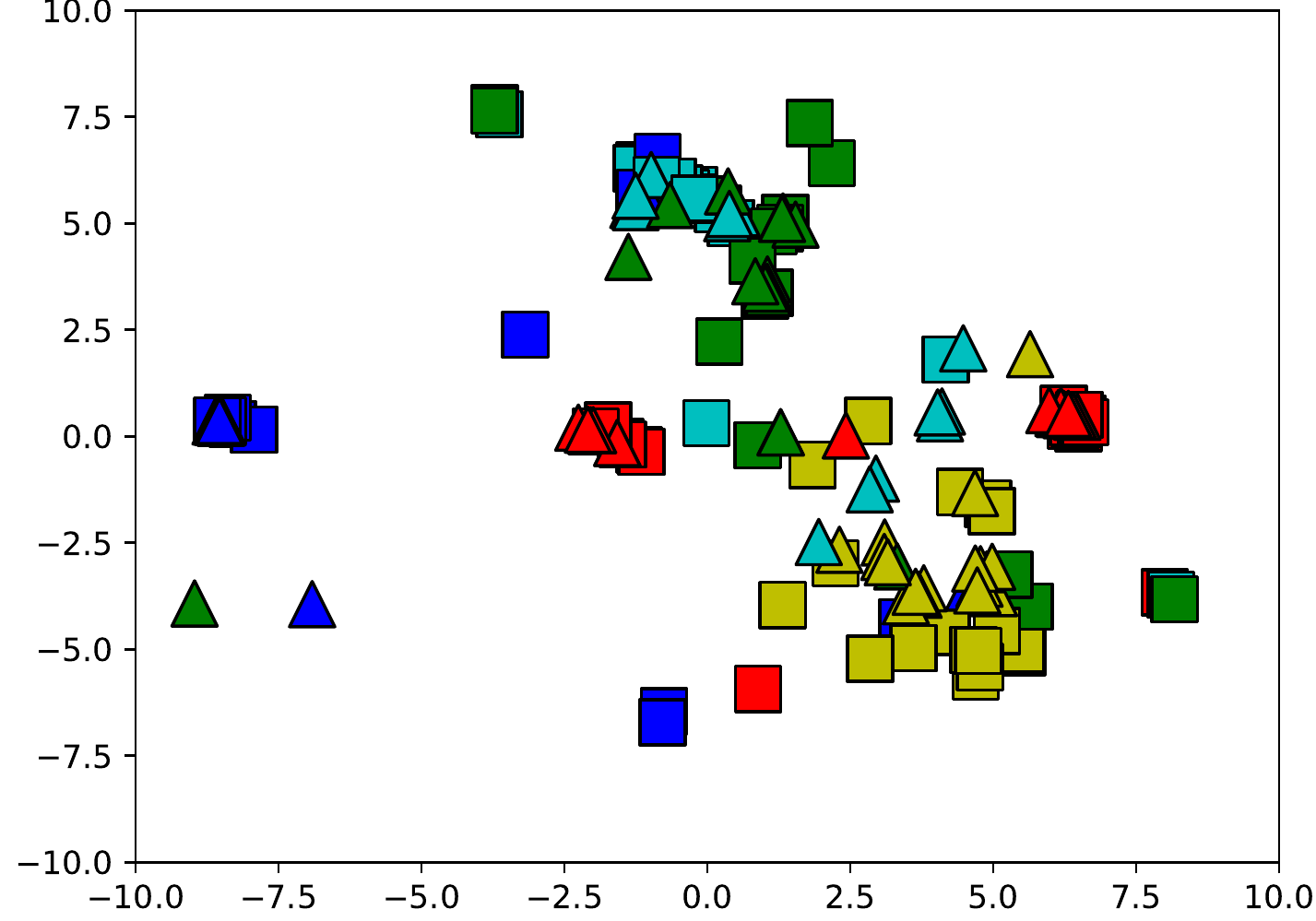}

		\label{fig:category_rgb_ske}
	\end{subfigure}	
    \begin{subfigure}[t]{0.24\linewidth}
		\centering\includegraphics[width=1.0\textwidth]{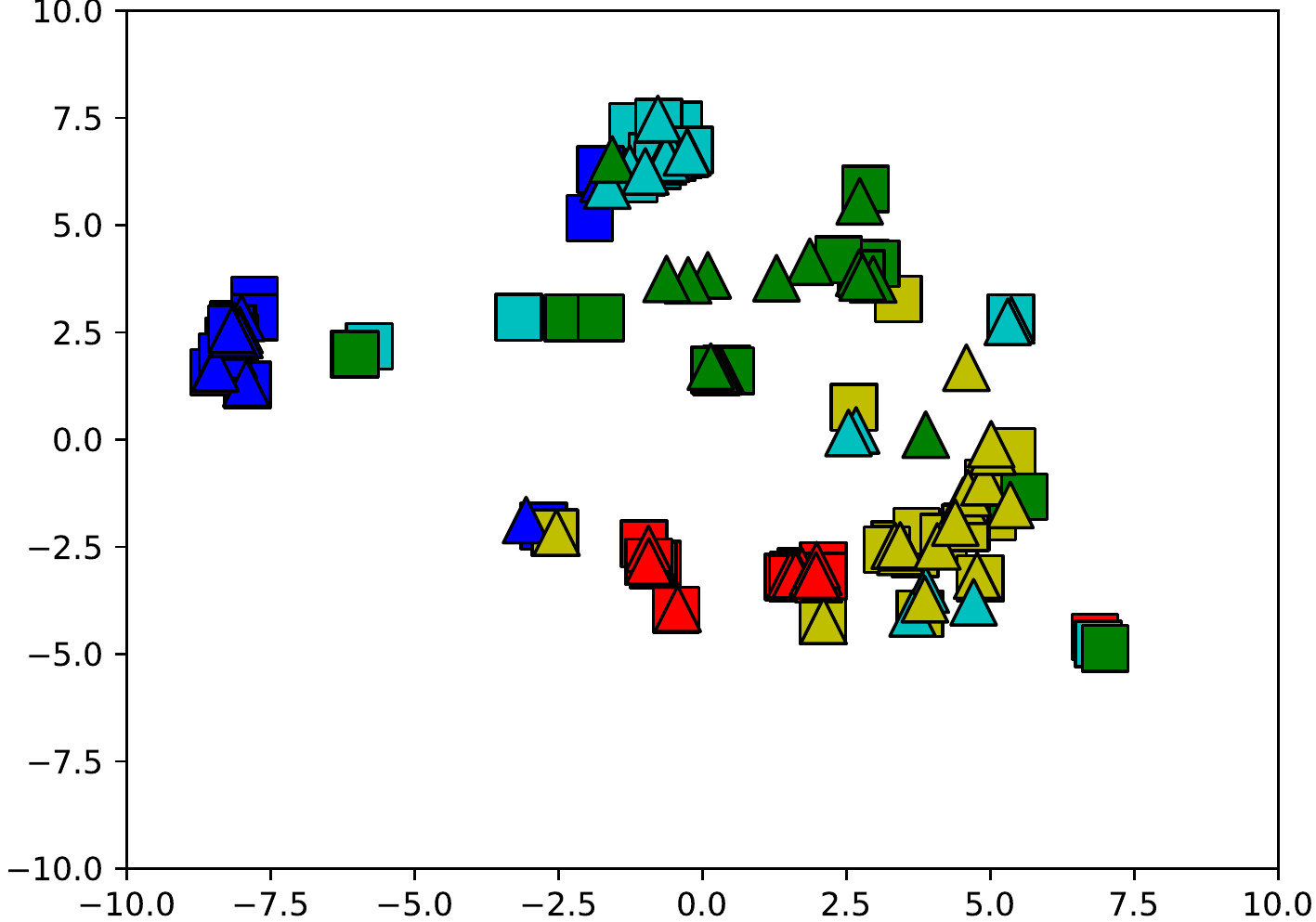}
				
		\label{fig:sample_rgb_ske}
	\end{subfigure}

	\begin{subfigure}[t]{0.24\linewidth}
		\centering\includegraphics[width=1.0\textwidth]{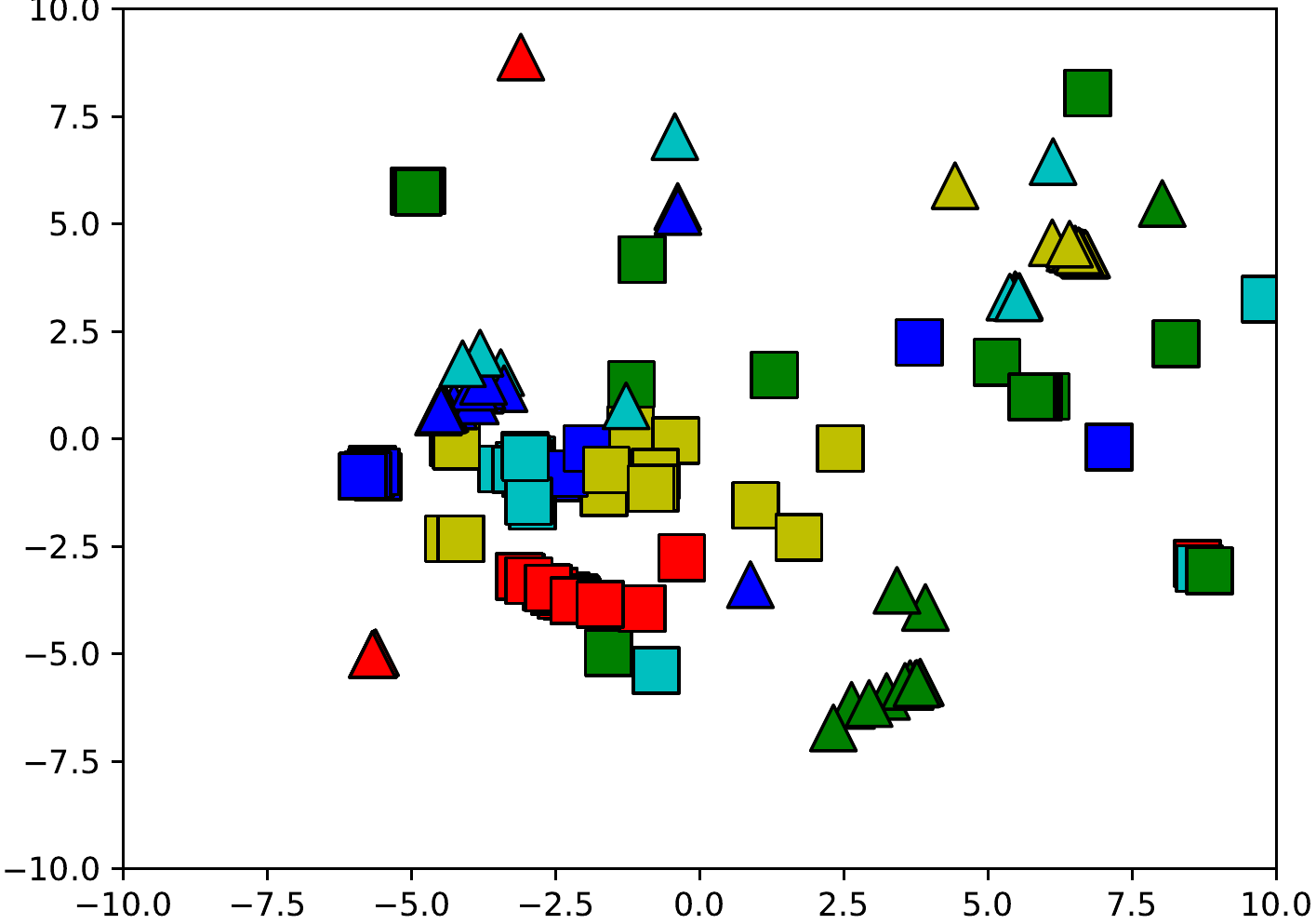}
		\caption{W/o adaptation}			
		\label{fig:baseline_opt_ske}
	\end{subfigure}
	\begin{subfigure}[t]{0.24\linewidth}
		\centering\includegraphics[width=1.0\textwidth]{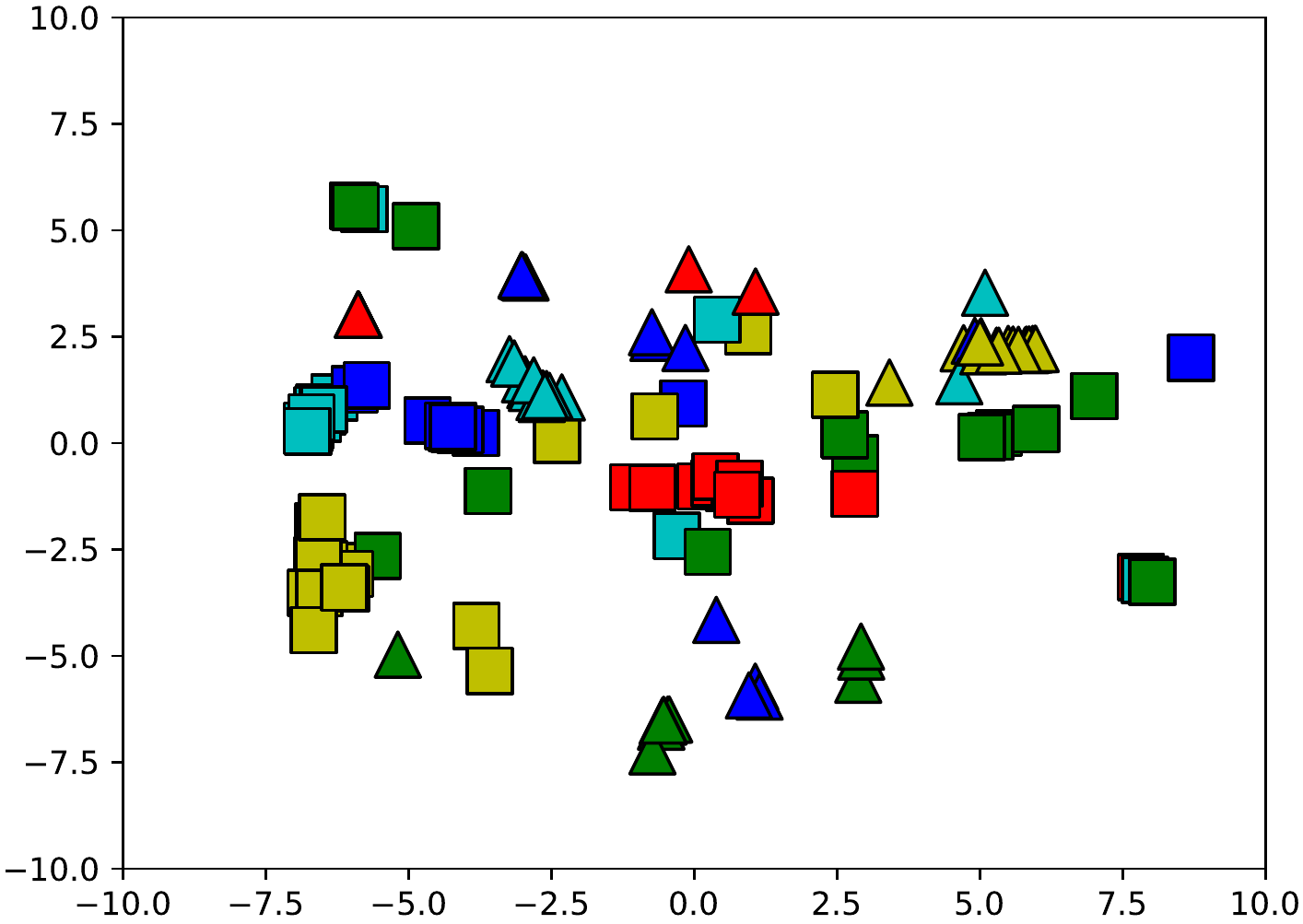}
		\caption{Domain-level}
		\label{fig:domain_opt_ske}
	\end{subfigure}	
    \begin{subfigure}[t]{0.24\linewidth}
    \centering\includegraphics[width=1.0\textwidth]{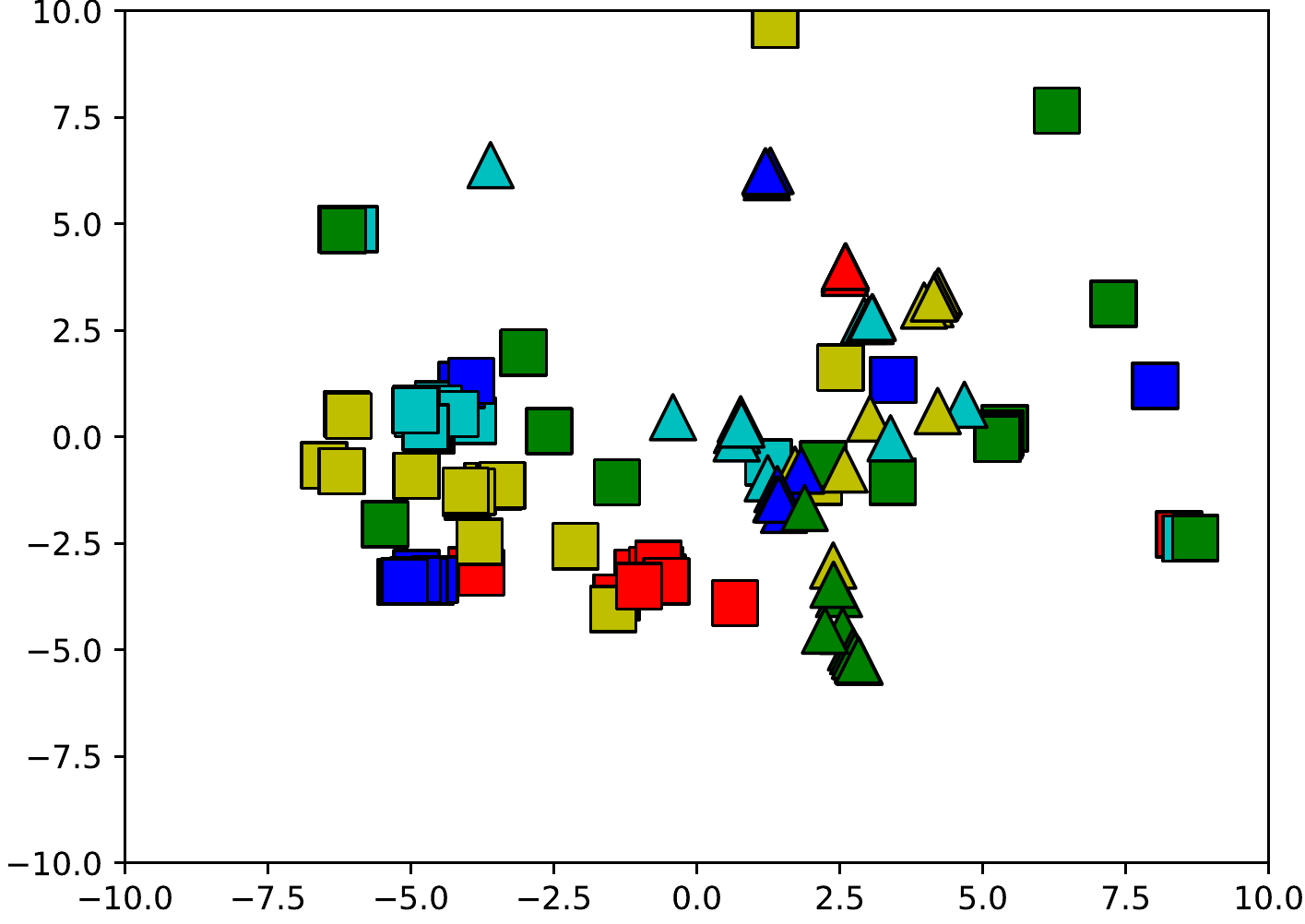}
		\caption{Category-level}
		\label{fig:category_opt_ske}
	\end{subfigure}	
    \begin{subfigure}[t]{0.24\linewidth}
		\centering\includegraphics[width=1.0\textwidth]{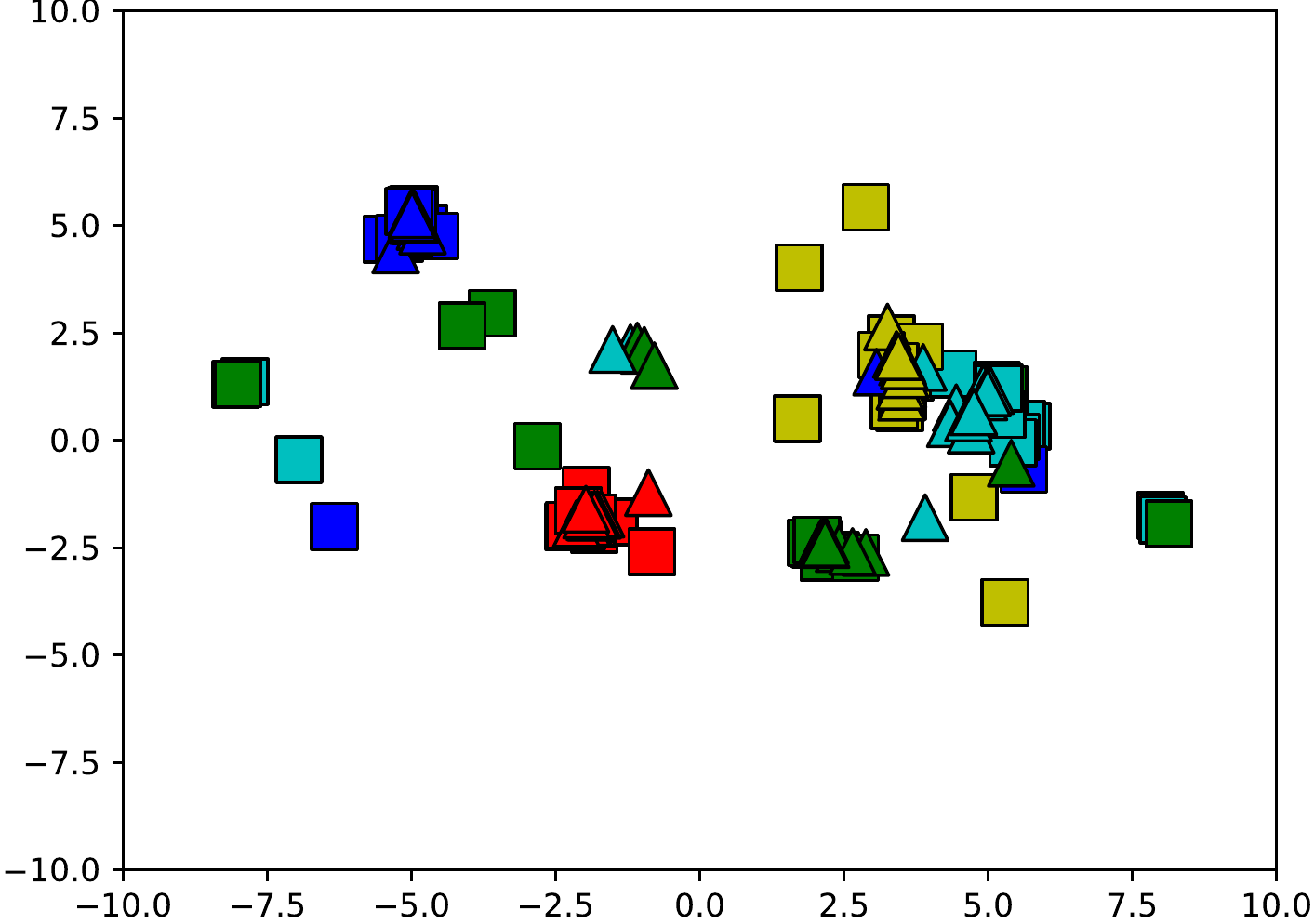}
		\caption{Sample-level}			
		\label{fig:sample_opt_ske}
	\end{subfigure}
	\caption{\M{The t-SNE results of source and auxiliary features with different modality adaptation levels (top: RGB \& skeletons, bottom: optical flow \& skeletons). We use the adapted features from the residual blocks in the Res-LSTM, D-Res-LSTM, C-Res-LSTM and S-Res-LSTM, respectively. Five classes are randomly selected from the testing data of the NTU (CV) dataset for visualization. Different colors denote different action classes. The features from source modalities are represented as triangles while those from skeletons are represented as squares.}}
\label{fig:tsne}
\end{figure*}

\R{\textbf{Parameter Analysis.} The hyper parameter $\lambda$ in Eq.~(\ref{equ:loss}) controls the degree of adaptive feature learning. To choose the scalar $\lambda$, we first observe the order of each item in the loss function, from which we roughly get the range of $\lambda$, then we vary different values to analyze its effect. The higher $\lambda$ forces the network to narrow the distance between the source and auxiliary modalities as close as possible. However, due to the noises in the skeleton data, the features from the source modality could be interrupted and thus influence the performance. Moreover, with a lower $\lambda$, the network only borrows little complementary information from additional skeletons. Therefore, we study the effect of $\lambda$ in Fig.~\ref{fig:effect_lambda} to seek the optimal. Note that $\lambda=0$ indicates the Res-LSTM results when no modality adaptation is adopted, which is marked by horizontal black dash lines. We show the performance on validation and test sets with various $\lambda$ in different levels of modality adaptation on the NTU (CS) dataset, respectively. In practice, the optimal $\lambda$ is selected from the best results on the validation set. However, as shown in Fig.~\ref{fig:effect_lambda}, different weights do not significantly influence the results. With $\lambda$ in the proper range, different levels of adaptation bring improvement on the results consistently. }

\textbf{Classification Analysis.}
In Fig.~\ref{fig:classification}, we list the histograms of the gain of S-Res-LSTM with respect to Res-LSTM. The gain values are calculated based on average precision for each class in the NTU dataset under the cross view setting. It is observed that for most classes, S-Res-LSTM outperforms the baseline. \M{In the meantime, for RGB, we find the actions related with motions are improved much more, such as \emph{clapping} (\#10), \emph{putting the palms together} (\#39), \emph{saluting} (\#38). With the auxiliary skeletons, our MCN is able to compensate source RGB features with motion information, increasing the discriminations of the actions. Meanwhile, the main contribution of skeleton data to optical flow is to provide extra posture information in actions where static frames are predominant, such as \emph{pointing to something with finger} (\#31), \emph{nausea} (\#48) and \emph{writing} (\#12). These actions with subtle optical flow are improved significantly with posture information compensated.}

\M{\textbf{Visualizations of Adapted Features.}}
To better understand the proposed modality adaptation method, we visualize the source and auxiliary feature vectors ($\mathbf{\hat{r}}_i$ and $\mathbf{\hat{a}}_i$ in Section~\ref{sec:proposed}) that are learned from different modality adaptation levels. The visualizations are achieved with t-SNE techniques, as shown in Fig.~\ref{fig:tsne}. We randomly select five classes from the testing data in the NTU (CV) dataset. We denote different action classes with different colors. The features from source modalities are represented with triangles and those from skeletons are represented with squares. The source modality of the figures in the first row is RGB and the second row is optical flow. It is noticed that with a finer adaptation level, the features from source and auxiliary modalities become ``closer" in the feature space. Compared with the distributions without modality adaptation, the source and auxiliary features after domain- or category-level adaptation stay tighter in general. However, when sample-level adaptation is adopted, the features in different modalities but sharing the same action class get together, which is able to further increase the discriminative performance.

\vspace{-2mm}
\section{Conclusion}
\label{sec:conclusion}
In this paper, we present a novel model for action recognition, Modality Compensation Network (MCN). Taking advantage of the auxiliary modality, we aim to compensate the feature learning in source modalities by adaptive representation learning. The modality adaptation block is developed to borrow the complementary information from the auxiliary modality, by narrowing the distance of source and auxiliary modal distributions. Adaptation schemes in different levels are employed according to the alignment of the training data. The ablation study illustrates the effectiveness of each component in our proposed network. Comprehensive analysis is given to better understand the modality adaptation scheme. Experiments demonstrate our model consistently improves the action recognition performance on four different datasets, and achieves remarkable performance compared with other state-of-the-arts.

\ifCLASSOPTIONcaptionsoff
  \newpage
\fi

\bibliographystyle{IEEEtran}
\bibliography{MCN_camera_ready}


\end{document}